\documentclass{article}

\usepackage[preprint,nonatbib]{neurips_2026}

\usepackage[utf8]{inputenc} % allow utf-8 input
\usepackage[T1]{fontenc}    % use 8-bit T1 fonts
\usepackage{hyperref}       % hyperlinks
\usepackage{url}            % simple URL typesetting
\usepackage{booktabs}       % professional-quality tables
\usepackage{amsfonts}       % blackboard math symbols
\usepackage{nicefrac}       % compact symbols for 1/2, etc.
\usepackage{microtype}      % microtypography
\usepackage[svgnames,dvipsnames]{xcolor}         % colors
\usepackage{colortbl}
\newcommand{\blue}[1]{\textcolor{blue}{#1}}

\newcommand{\gray}[1]{\textcolor{gray!85}{#1}}
\newcolumntype{T}{>{\columncolor{gray!15}}c}
\usepackage{graphicx}
\usepackage{wrapfig}
\usepackage{bm}
\usepackage{amsmath}
\usepackage{textcomp}
\usepackage{array}
\usepackage{tabularx}
\usepackage{makecell}
\usepackage{multirow}
\usepackage{csquotes}
\usepackage{bbding}
\usepackage{pifont}
\usepackage{subcaption}
\usepackage{algorithm}
\usepackage{algpseudocode}
\usepackage{microtype}
\usepackage{hyphenat}
\usepackage{floatflt}
\usepackage[capitalize]{cleveref}
\usepackage{mathtools}
\usepackage{fancyvrb}

\DefineVerbatimEnvironment{promptbox}{Verbatim}{
  fontsize=\scriptsize,
  frame=single,
  framerule=0.5pt,
  framesep=2mm,
  rulecolor=\color{gray!50},
}

\title{DASH-OPD: Discrepancy-Aware Switching with Hysteresis for On-Policy Distillation}

\author{%
  Yuchen Xia\textsuperscript{1,2}
  \quad
  Qianguo Sun\textsuperscript{2}
  \quad
  Chao Song\textsuperscript{2}
  \quad
  Junlong Wu\textsuperscript{3}
  \quad
  Yiyan Qi\textsuperscript{2}\thanks{Corresponding authors}
  \quad
  Yunjian Xu\textsuperscript{1}\footnotemark[\value{footnote}]\\
  \textsuperscript{1}The Chinese University of Hong Kong
  \quad
  \textsuperscript{2}IDEA Research
  \quad
  \textsuperscript{3}Emdoor Research Institute\\
  \textsuperscript{1}\texttt{\{ycxia,yjxu\}@mae.cuhk.edu.hk}\\
  \quad
  \textsuperscript{2}\texttt{\{xiayuchen,sunqianguo,songchao,qiyiyan\}@idea.edu.cn}\\
  \quad
  \textsuperscript{3}\texttt{junlong.wu@emdoor.com}
}

\begin{document}

\maketitle

\begin{abstract}
While on-policy distillation (OPD) reduces exposure bias by training student language models on their own rollouts, early student errors in long-horizon agentic scenarios can lead to contexts unfamiliar to the teacher. To improve trajectory quality, recent work on agentic OPD introduces teacher intervention into training rollouts by switching the executor between the student and the teacher. However, existing methods determine how much teacher intervention is needed---but not when. To address this limitation, we propose DASH-OPD (Discrepancy-Aware Switching with Hysteresis for OPD), the first agentic OPD method to perform adaptive, bidirectional executor switching. At each turn, DASH-OPD measures teacher--student discrepancy using a mean log-likelihood ratio over action tokens. High student-to-teacher ratios on student turns serve as drift signals, while low teacher-to-student ratios on teacher turns serve as recovery signals. These signals are accumulated over multiple turns to form drift and recovery evidence, respectively. DASH-OPD switches executors when either type of evidence exceeds its corresponding switching threshold, introducing hysteresis that prevents rapid switching triggered by transient discrepancy fluctuations. Across three benchmarks and two student model scales, DASH-OPD outperforms five baselines in all 14 task-performance comparisons, while requiring the fewest interaction turns in nine of ten efficiency comparisons. Code, trained models, and training logs are available at \url{https://github.com/Lucian1115/DASH-OPD}.
\end{abstract}

\section{Introduction}

Deploying capable large language models (LLMs) can be computationally expensive, motivating knowledge distillation to transfer the capabilities from teacher models to more efficient student models \cite{kd,kd_llm}. Conventional distillation trains the student on teacher-generated \cite{teacher-generated} or human-provided \cite{human-provided} trajectory data, whereas during inference the student must condition on self-generated contexts, thereby creating a training--inference mismatch \cite{dagger}, known as exposure bias \cite{exposure_bias}. On-policy distillation (OPD) \cite{opd} mitigates exposure bias by training students on self-generated trajectories \cite{opd_survey1,opd_survey2}. However, this benefit can diminish in long-horizon agentic tasks: accumulated student errors steer trajectories into contexts unfamiliar to the teacher \cite{sage-opd}, thereby degrading the reliability of teacher supervision \cite{sod,fu2026revisiting,opd_survey3}. In our experiments, vanilla OPD marginally improves the 4B student's success rate on the ScienceWorld benchmark \cite{scienceworld}, from 16.21\% to 16.36\% (see \cref{tab:main-results-all-envs}).

\begin{wrapfigure}{r}{0.45\textwidth}
  \vspace{-1em}
  \centering
  \includegraphics[width=\linewidth]{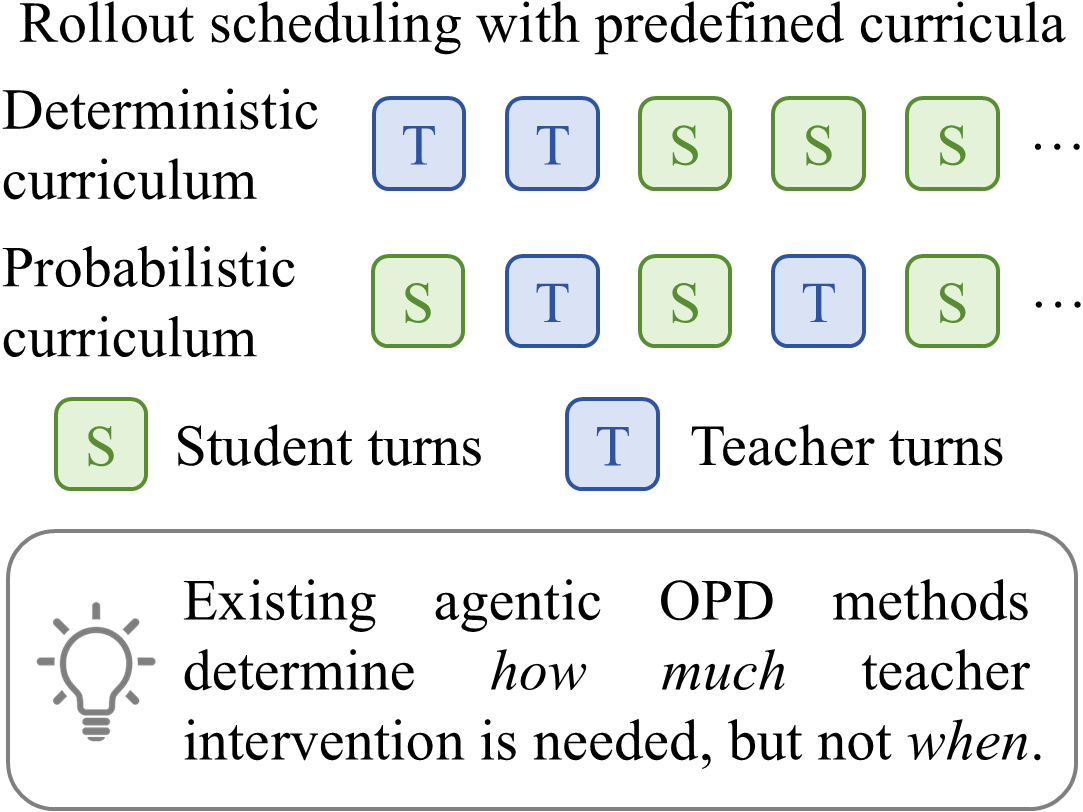}
  \caption{A research gap in agentic OPD.}
  \label{fig:existing-methods}
\end{wrapfigure}

Teacher-generated turns can recover trajectories from student-induced drift, but excessive teacher intervention reintroduces exposure bias. Recent work on agentic OPD uses predefined curricula \cite{curriculum} to balance teacher and student control, as illustrated in \cref{fig:existing-methods}. TCOD \cite{tcod} follows a deterministic curriculum that progressively lengthens a contiguous student-controlled segment in each trajectory. Guided-OPD \cite{guided-opd} follows a probabilistic curriculum to sample the executor at each turn, with the probability of selecting the teacher gradually decreasing over training. However, existing methods determine \emph{how much} teacher intervention is needed, but not \emph{when}, even though its necessity may vary within and across trajectories \cite{gao2026agentcollab,cai2025robot,thriftydagger}.

Teacher--student discrepancy can indicate when the teacher should intervene and when the student can resume control. We therefore propose DASH-OPD (Discrepancy-Aware Switching with
Hysteresis for OPD), as illustrated in \cref{fig:dash-overview}. At each turn, either the student or the teacher generates a response, and the other model scores the action tokens in this response under the same context, yielding a mean log-likelihood ratio that quantifies teacher--student discrepancy. On student turns, high student-to-teacher ratios serve as drift signals. On teacher turns, low teacher-to-student ratios serve as recovery signals. These two types of signals are accumulated separately across turns to form drift and recovery evidence. Drift evidence tracks the need for teacher intervention, while recovery evidence tracks whether the student can resume control. DASH-OPD switches control to the teacher when drift evidence exceeds an intervention threshold and back to the student when recovery evidence exceeds a return threshold. Following Guided-OPD, DASH-OPD trains the student on all response tokens using reverse-KL distillation on student turns and forward-KL distillation on teacher turns.

DASH-OPD achieves hysteretic switching by acting on evidence accumulated across multiple turns rather than on individual signals. Requiring sufficient evidence before switching makes DASH-OPD robust to transient fluctuations in teacher--student discrepancy, thereby preventing rapid oscillation between the two models.

To the best of our knowledge, DASH-OPD is the \emph{first} agentic OPD method to perform adaptive, bidirectional executor switching. This mechanism enables DASH-OPD to invoke teacher intervention when needed. We evaluate DASH-OPD with 1.7B and 4B student models on WebShop \cite{webshop}, ALFWorld \cite{alfworld}, and ScienceWorld \cite{scienceworld}. DASH-OPD outperforms five baselines in all 14 task-performance comparisons, while requiring the fewest interaction turns in nine of ten efficiency comparisons (second fewest on ScienceWorld at 1.7B).

\section{Related work}

To balance trajectory quality with the student's exposure to self-generated contexts during training, recent OPD methods schedule rollouts at either turn or token granularity. Across both granularities, some methods include teacher-generated segments in training rollouts, whereas others keep the rollouts entirely student-generated.

\paragraph{Turn-level rollout scheduling.}
In agentic OPD, each turn forms a complete decision unit, making turn-level rollout scheduling a natural choice. TCOD \cite{tcod} and Guided-OPD \cite{guided-opd} both include teacher-generated turns and regulate intervention through predefined curricula. TCOD follows a deterministic curriculum that gradually expands the student rollout horizon. Guided-OPD follows a probabilistic curriculum that samples either the teacher or the student to act at each turn, with the probability of choosing the teacher gradually decreasing over training. TurnOPD \cite{turnopd} uses student-only rollouts and adapts the rollout horizon based on turn-level statistics.

\paragraph{Token-level rollout scheduling.}
For general text-generation tasks such as coding and mathematical reasoning, rollout scheduling naturally operates at the token level. AdaSwitch \cite{adaswitch} uses a context-dependent threshold to trigger a student-to-teacher switch, after which the teacher completes the response. POPD and TOPD \cite{popd_topd} both use student-generated trajectories without teacher intervention. They implement two forms of rollout scheduling: POPD follows a predetermined curriculum that progressively expands the student rollout horizon over training, whereas TOPD keeps the horizon fixed. ESR \cite{esr} likewise limits each student rollout to a fixed number of tokens.

\paragraph{\normalfont However, it remains an open challenge to determine when the teacher should intervene and when the student should resume control.}

\section{Preliminaries}

\begin{figure}[t]
    \centering
    \includegraphics[width=\linewidth]{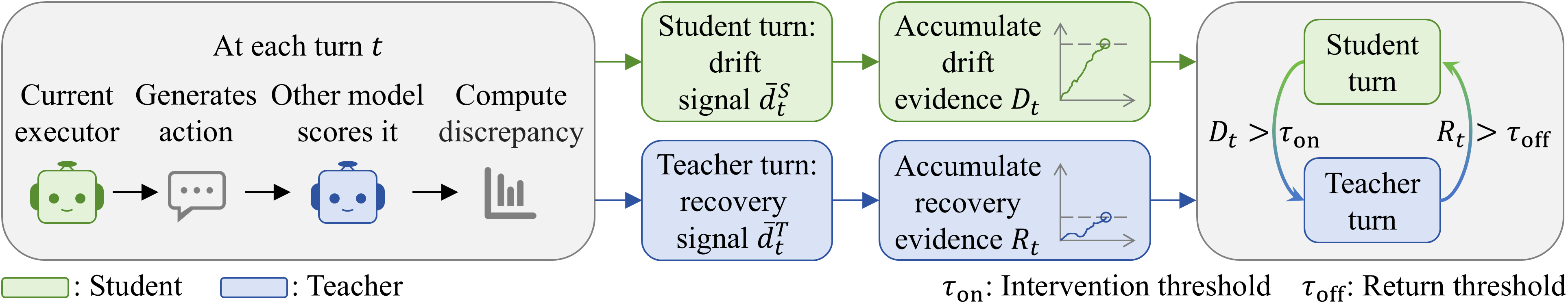}
    \caption{Pipeline of DASH-OPD's switching mechanism. DASH-OPD adaptively switches control between the student and teacher based on their discrepancy.}
    \label{fig:dash-overview}
\end{figure}

\paragraph{Large language model (LLM) agents.}

Rather than generating a single response to a fixed prompt, an LLM agent interacts with an environment over multiple turns \cite{react,liu2024agentbench}. At turn $t$, the agent conditions its response $y_t$ on the interaction history $h_t=(x,o_1,y_1,\ldots,o_t)$, where $x$ denotes the task prompt and $o_t$ denotes the agent's observation. Executing the action parsed from $y_t$ yields the next observation $o_{t+1}$. The response $y_t$ is a token sequence $(y_{t,1},\ldots,y_{t,L_t})$, where $L_t$ is the sequence length. The agent is modeled as a conditional policy $\pi(y_t\mid h_t)$:
\begin{equation}
    \pi(y_t\mid h_t)=\prod_{i=1}^{L_t}
    \pi(y_{t,i}\mid h_t,y_{t,<i}),
    \quad
    y_{t,<i}=(y_{t,1},\ldots,y_{t,i-1}).
    \label{eq:response-policy}
\end{equation}

\paragraph{On-policy distillation (OPD).}

Conventional knowledge distillation trains the student on trajectories provided by the teacher \cite{teacher-generated} or humans \cite{human-provided}. During inference, however, the student must condition on self-generated contexts. This training--inference mismatch, known as exposure bias \cite{exposure_bias}, limits the student's ability to recover from its own errors.

OPD mitigates exposure bias by performing distillation on student-generated trajectories \cite{opd}. Given a trainable student policy $\pi_\theta$ and a frozen teacher policy $\pi_T$, two common distillation objectives are the forward- and reverse-KL divergences between the two policies. The forward-KL objective is:
\begin{equation}
    \mathcal L_{\mathrm{forward}}
    =\sum_t\mathbb E_{h_t\sim \rho_{\pi_\theta}^t}
      D_{\mathrm{KL}}\!\left(
      \pi_T(\cdot\mid h_t)\,\|\,\pi_\theta(\cdot\mid h_t)
      \right).
\label{eq:forward-KL}
\end{equation}
The reverse-KL objective is:
\begin{equation}
    \mathcal L_{\mathrm{reverse}}
    =\sum_t\mathbb E_{h_t\sim \rho_{\pi_\theta}^t}
      D_{\mathrm{KL}}\!\left(
      \pi_\theta(\cdot\mid h_t)\,\|\,\pi_T(\cdot\mid h_t)
      \right).
\label{eq:reverse-KL}
\end{equation}
Here, $\rho_{\pi_\theta}^t$ denotes the student-induced context distribution at turn $t$. In practice, training trajectories are generated by a frozen student snapshot $\pi_{\bar\theta}$ to keep the rollout distribution fixed while $\pi_\theta$ is updated. The forward-KL objective is mode-covering, encouraging the student to capture diverse teacher behaviors, whereas the reverse-KL objective is mode-seeking, concentrating the student on high-probability teacher behaviors. At inference time, only the student policy is deployed.

\section{Method}

\subsection{Overview}

At each turn $t$, DASH-OPD selects either the student or the teacher to generate the response. Let $m_t\in\{S,T\}$ ($S$ for student and $T$ for teacher) indicate which model controls turn $t$. The response $y_t$ is generated as follows:
\begin{equation}
    y_t\sim
    \begin{cases}
        \pi_{\bar\theta}(\cdot\mid h_t), & m_t=S,\\
        \pi_T(\cdot\mid h_t), & m_t=T.
    \end{cases}
    \label{eq:dash-generation}
\end{equation}
The non-executing model scores the action tokens in $y_t$ to compute a directional teacher--student discrepancy. DASH-OPD switches control between the two models based on the evidence accumulated from these discrepancies. The pipeline of DASH-OPD's switching mechanism is illustrated in \cref{fig:dash-overview}. Appendix~\ref{app:algorithm} provides the pseudocode for DASH-OPD.

\subsection{Directional teacher--student discrepancy}

If turn $t$ is student-controlled, the teacher measures the discrepancy by computing a student-to-teacher mean log-likelihood ratio (i.e., a sample-based analogue of $D_{\mathrm{KL}}(\pi_{\bar\theta}\Vert\pi_T)$):
\begin{equation}
d_t^S=\frac{1}{N_t^A}\sum_i A_{t,i}
  [\log\pi_{\bar\theta}(y_{t,i}\mid h_t,y_{t,<i})
  -\log\pi_T(y_{t,i}\mid h_t,y_{t,<i})], \quad
  y_t\sim\pi_{\bar\theta},
\label{eq:student-ratio}
\end{equation}
where $A_{t,i}$ is an action mask ($1$ for tokens in the action content; $0$ otherwise), and $N_t^A=\sum_i A_{t,i}$ is the
number of action tokens. If turn $t$ is teacher-controlled, the student measures the discrepancy in the opposite direction by computing a teacher-to-student mean log-likelihood ratio (i.e., a sample-based analogue of $D_{\mathrm{KL}}(\pi_T\Vert\pi_{\bar\theta})$):
\begin{equation}
d_t^T=\frac{1}{N_t^A}\sum_i A_{t,i}
  [\log\pi_T(y_{t,i}\mid h_t,y_{t,<i})
  -\log\pi_{\bar\theta}(y_{t,i}\mid h_t,y_{t,<i})], \quad
  y_t\sim\pi_T.
\label{eq:teacher-ratio}
\end{equation}

The teacher--student discrepancy is measured only over the action portion of the response, focusing on decision-making rather than reasoning patterns (see Appendix~\ref{app:action-extraction-robustness} for robustness to action extraction failures). Both $d_t^S$ and $d_t^T$ are directly computed from the log-probabilities collected during rollout scoring, requiring no further model forward passes for switching. Appendix~\ref{app:directional-discrepancy} provides a detailed rationale for the directionality of teacher--student discrepancy.

\subsection{From drift and recovery signals to evidence}

DASH-OPD detects a drift signal when $d_t^S$ rises above its historical level, and detects a recovery signal when $d_t^T$ falls below its historical level. To track these historical levels, DASH-OPD maintains running statistics $\mathcal R_t^S=(\mu_t^S,v_t^S)$ for $d_t^S$ and $\mathcal R_t^T=(\mu_t^T,v_t^T)$ for $d_t^T$, updated on student and teacher turns, respectively. Here, $\mu_t^{m_t}$ and $v_t^{m_t}$ denote the mean and variance of observed $d_t^{m_t}$. These statistics are retained throughout each trajectory without resetting. To obtain drift and recovery signals, DASH-OPD standardizes $d_t^{m_t}$ using the corresponding statistics:
\begin{equation}
    \bar{d}_t^{m_t} =
    \frac{d_t^{m_t}-\mu_{t-1}^{m_t}}
         {\sqrt{v_{t-1}^{m_t}}}, \quad
    m_t\in\{S,T\}.
    \label{eq:dash-standardization}
\end{equation}
Here, $\bar{d}_t^S$ denotes the drift signal, and $\bar{d}_t^T$ denotes the recovery signal. On the first turn of each executor in a trajectory, $\bar{d}_t^{m_t}$ is set to zero.

DASH-OPD accumulates $\bar{d}_t^S$ and $\bar{d}_t^T$ separately across turns to form drift evidence $D_t$ and recovery evidence $R_t$. On student turns, DASH-OPD updates $D_t$ as follows:
\begin{equation}
    D_t = \max(D_{t-1}+\bar{d}_t^S, 0), \quad
         m_t=S.
\label{eq:drift-evidence}
\end{equation}
On teacher turns, DASH-OPD updates $R_t$ as follows:
\begin{equation}
    R_t = \max(R_{t-1}-\bar{d}_t^T, 0), \quad
         m_t=T.
\label{eq:recovery-evidence}
\end{equation}
Both $D_t$ and $R_t$ are initialized to zero for each trajectory.
Positive $\bar{d}_t^S$ increases $D_t$, whereas negative $\bar{d}_t^T$ increases $R_t$. Signals in the opposite directions reduce the respective evidence toward zero.

\subsection{Switching mechanism}

DASH-OPD switches control to the teacher when the drift evidence $D_t$ exceeds an intervention threshold $\tau_{\mathrm{on}}$, and returns control to the student when the recovery evidence $R_t$ exceeds a return threshold $\tau_{\mathrm{off}}$.
To account for differences in rollout horizon across benchmarks, we scale both thresholds by $H_{\max}$, the maximum number of turns per trajectory:
\begin{equation}
    \tau_{\mathrm{on}} = q_{\mathrm{on}} H_{\max}, \quad
    \tau_{\mathrm{off}} = q_{\mathrm{off}} H_{\max},
    \label{eq:dash-thresholds}
\end{equation}
where $q_{\mathrm{on}}$ and $q_{\mathrm{off}}$ are the intervention and return threshold ratios, respectively.
In addition to the evidence-based switching, DASH-OPD also triggers teacher intervention when the student stagnates and enforces a minimum length of teacher-controlled segments.

Specifically, after a student turn, the condition for teacher intervention is:
\begin{equation}
\mathrm{intervention}_t
  = (D_t>\tau_{\mathrm{on}}) \lor g_t.
\label{eq:support-switching}
\end{equation}
Here, $g_t$ denotes a stagnation flag that is activated when the same action is taken or the same observation is encountered for $K$ consecutive student turns. 
If $\mathrm{intervention}_t$ holds, DASH-OPD sets $m_{t+1}=T$ and resets the drift evidence $D_t$ to zero.
After a teacher turn, the condition for returning control to the student is:
\begin{equation}
\mathrm{return}_t
  = (R_t>\tau_{\mathrm{off}}) \land (\ell_t\ge \ell_{\min}).
\label{eq:return-switching}
\end{equation}
Here, $\ell_t$ denotes the length of the current teacher-controlled segment, and $\ell_{\min}$ specifies the minimum length for teacher segments. If $\mathrm{return}_t$ holds, DASH-OPD sets $m_{t+1}=S$ and resets the recovery evidence $R_t$ to zero. See Appendix~\ref{app:sudden-difficulty} for handling sudden increases in task difficulty.

The switching threshold ratios $q_{\mathrm{on}}$ and $q_{\mathrm{off}}$,
the minimum teacher segment length $\ell_{\min}$, and the stagnation
threshold $K$ are held fixed across all
benchmarks and student model scales.

\paragraph{Rollout initialization.}
For each trajectory, DASH-OPD randomly selects either the student or the teacher to control the first turn. The teacher-start probability decreases linearly as training progresses:
\begin{equation}
    p_{T}(s)
    =1-s/S_{\max},
    \label{eq:teacher-start}
\end{equation}
where $s$ is the current training step and $S_{\max}$ is the total number of training steps. This initialization encourages more teacher intervention early in training and gradually reduces it as the student improves.

\subsection{Distillation objective}

Following Guided-OPD \cite{guided-opd}, we train the student using sampled reverse-KL
distillation on student-controlled turns and sampled forward-KL distillation on
teacher-controlled turns. Let $\mathcal L_{t,i}^{S}$ and $\mathcal L_{t,i}^{T}$
denote the respective training losses for token $i$ at turn $t$ (detailed in Appendix~\ref{app:token-losses}), and define
$\mathbb I_t^S=\mathbb I[m_t=S]$ and $\mathbb I_t^T=\mathbb I[m_t=T]$.
The overall objective is:
\begin{equation}
    \mathcal L_{\mathrm{overall}}
    =
    \frac{\sum_{t,i}M_{t,i}
    \left(\mathbb I_t^S\mathcal L_{t,i}^S
    +\mathbb I_t^T\mathcal L_{t,i}^T\right)}
    {\max\!\left(\sum_{t,i}M_{t,i},1\right)} ,
    \label{eq:dash-objective}
\end{equation}
where $M_{t,i}$ equals $1$ for valid response tokens and $0$ otherwise.
The distillation is applied to the full response, including both reasoning and action
tokens, whereas the executor switching relies on the action-only discrepancy.

\section{Experiments}

\begin{table}[t]
    \centering
    \caption{Task performance and deployment efficiency across three benchmarks and two student model scales. For ALFWorld, \enquote{Overall} reports
    results over all tasks in the \enquote{IID} and \enquote{OOD} test splits.
    Higher \enquote{Score} and \enquote{SR} indicate better task performance, while fewer \enquote{Turns} indicate greater efficiency.
    Results for the zero-shot teacher and students are shown for reference and excluded from the rankings.
    Within each model scale, \blue{\textbf{blue bold}} and
    \underline{underline} mark the best and second-best results,
    respectively (ties share the same style).
    DASH-OPD leads all 14 task-performance comparisons
    and requires the fewest turns in 9 of 10 efficiency comparisons (ranking second in the remaining one).}
    \label{tab:main-results-all-envs}
    \small
    \setlength{\tabcolsep}{2.1pt}
    \resizebox{\textwidth}{!}{%
    \begin{tabular}{@{}lccTcTcTcTccT@{}}
        \toprule
        \multirow{3}{*}{Method}
        & \multicolumn{3}{c}{WebShop}
        & \multicolumn{6}{c}{ALFWorld}
        & \multicolumn{3}{c}{ScienceWorld} \\
        \cmidrule(lr){2-4}\cmidrule(lr){5-10}\cmidrule(l){11-13}
        & \multicolumn{3}{c}{}
        & \multicolumn{2}{c}{IID} & \multicolumn{2}{c}{OOD}
        & \multicolumn{2}{c}{Overall} & \multicolumn{3}{c}{} \\
        \cmidrule(lr){5-6}\cmidrule(lr){7-8}\cmidrule(lr){9-10}
        & Score $\uparrow$ & SR $\uparrow$ & Turns $\downarrow$
        & SR $\uparrow$ & Turns $\downarrow$ & SR $\uparrow$ & Turns $\downarrow$
        & SR $\uparrow$ & Turns $\downarrow$
        & Score $\uparrow$ & SR $\uparrow$ & Turns $\downarrow$ \\
        \midrule
        \multicolumn{13}{@{}l}{Qwen3-30B-A3B teacher} \\
        \gray{Zero-shot}
        & \gray{40.67} & \gray{17.00\%} & \gray{8.06}
        & \gray{45.00\%} & \gray{23.13} & \gray{36.57\%} & \gray{24.27} & \gray{40.88\%} & \gray{23.69}
        & \gray{51.86} & \gray{23.01\%} & \gray{21.23} \\
        \midrule
        \multicolumn{13}{@{}l}{Qwen3-1.7B student} \\
        \gray{Zero-shot}
        & \gray{3.33} & \gray{2.00\%} & \gray{14.68}
        & \gray{2.86\%} & \gray{29.28} & \gray{0.00\%} & \gray{30.00} & \gray{1.46\%} & \gray{29.63}
        & \gray{10.92} & \gray{1.76\%} & \gray{17.01} \\
        Vanilla OPD
        & 32.03 & 10.00\% & 8.40
        & 23.57\% & 25.47 & 29.10\% & 25.01 & 26.28\% & 25.25
        & 27.23 & 7.65\% & 25.58 \\
        AdaSwitch
        & 34.43 & \underline{12.00\%} & 8.05
        & \underline{27.14\%} & 25.58 & 26.12\% & 25.54 & 26.64\% & 25.56
        & \underline{35.85} & \underline{11.01\%} & 23.96 \\
        TCOD
        & 36.41 & 11.00\% & \underline{7.68}
        & 25.71\% & 25.42 & \underline{32.09\%} & \underline{24.99} & 28.83\% & \underline{25.21}
        & 30.45 & 5.93\% & \blue{\textbf{22.80}} \\
        Guided-OPD
        & 33.19 & 10.00\% & 8.34
        & 26.43\% & 25.74 & 30.60\% & \underline{24.99} & 28.47\% & 25.37
        & 32.14 & 10.24\% & 25.23 \\
        TurnOPD
        & \underline{37.29} & \underline{12.00\%} & 7.96
        & \underline{27.14\%} & \underline{25.41} & 31.34\% & 25.01 & \underline{29.20\%} & 25.22
        & 32.52 & 9.33\% & 24.91 \\
        \textbf{DASH-OPD}
        & \blue{\textbf{43.53}} & \blue{\textbf{13.00\%}} & \blue{\textbf{6.25}}
        & \blue{\textbf{31.43\%}} & \blue{\textbf{24.98}} & \blue{\textbf{34.33\%}} & \blue{\textbf{24.69}}
        & \blue{\textbf{32.85\%}} & \blue{\textbf{24.84}}
        & \blue{\textbf{38.46}} & \blue{\textbf{12.77\%}} & \underline{23.60} \\
        \midrule
        \multicolumn{13}{@{}l}{Qwen3-4B student} \\
        \gray{Zero-shot}
        & \gray{22.74} & \gray{9.00\%} & \gray{11.41}
        & \gray{31.43\%} & \gray{24.49} & \gray{36.57\%} & \gray{23.83} & \gray{33.94\%} & \gray{24.16}
        & \gray{38.39} & \gray{16.21\%} & \gray{19.08} \\
        Vanilla OPD
        & 35.10 & 12.00\% & 8.44
        & 34.29\% & 24.38 & 38.81\% & 24.40 & 36.50\% & 24.39
        & 43.71 & 16.36\% & 23.19 \\
        AdaSwitch
        & 34.85 & 14.00\% & 8.36
        & 35.00\% & 23.75 & 38.06\% & 23.92 & 36.50\% & \underline{23.83}
        & 46.19 & \underline{20.34\%} & 21.36 \\
        TCOD
        & \underline{38.05} & \underline{14.50\%} & \underline{8.02}
        & \underline{37.86\%} & \underline{23.64} & 34.70\% & 24.20 & 36.31\% & 23.91
        & \underline{47.30} & 19.69\% & 21.98 \\
        Guided-OPD
        & 35.82 & 13.00\% & 8.42
        & 35.00\% & 23.79 & 32.09\% & 24.72 & 33.58\% & 24.24
        & 46.05 & 19.65\% & \underline{21.12} \\
        TurnOPD
        & 36.26 & 14.00\% & 8.24
        & 35.71\% & 23.88 & \underline{39.55\%} & \underline{23.88} & \underline{37.59\%} & 23.88
        & 45.26 & 17.20\% & 22.45 \\
        \textbf{DASH-OPD}
        & \blue{\textbf{42.77}} & \blue{\textbf{16.00\%}} & \blue{\textbf{7.15}}
        & \blue{\textbf{41.43\%}} & \blue{\textbf{22.86}} & \blue{\textbf{41.04\%}} & \blue{\textbf{23.46}}
        & \blue{\textbf{41.24\%}} & \blue{\textbf{23.15}}
        & \blue{\textbf{50.97}} & \blue{\textbf{22.40\%}} & \blue{\textbf{20.62}} \\
        \bottomrule
    \end{tabular}%
    }
\end{table}

\subsection{Experimental setup}
\label{sec:experimental-setup}

\paragraph{Baseline methods.}
We compare DASH-OPD against five baselines: vanilla OPD
\cite{opd}; a general-purpose OPD method,
AdaSwitch \cite{adaswitch}; and three agentic OPD methods, TCOD
\cite{tcod}, Guided-OPD \cite{guided-opd}, and TurnOPD \cite{turnopd}.
For TCOD, we report the average performance of its B2F and F2B variants (individual results are provided in Appendix~\ref{app:tcod-variants}).
We adapt AdaSwitch to agentic OPD by applying its token-level
switching rule independently within each turn.
We exclude POPD/TOPD \cite{popd_topd} and ESR \cite{esr},
as their token-level rollout scheduling may truncate a turn
before its final action block. We distill a frozen Qwen3-30B-A3B teacher into Qwen3-1.7B and Qwen3-4B students \cite{qwen3}. The zero-shot results for the teacher and student models are reported for reference but excluded from the rankings. To ensure a fair comparison, all baselines and DASH-OPD use the same settings
for shared training and evaluation components
(see \cref{tab:shared-configurations} in Appendix~\ref{app:full-configurations}). For most settings specific to each baseline, we follow the original papers or official implementations
(see \cref{tab:baseline-configurations} in Appendix~\ref{app:full-configurations}). For the few baseline settings that require adaptation to our experimental setup, we use the same tuning budget as for DASH-OPD.

\paragraph{Benchmarks and metrics.}
We evaluate all methods on three benchmarks for long-horizon agentic tasks with text-based interactions:
WebShop \cite{webshop}, which requires searching for and selecting products on a simulated shopping website to satisfy user requirements;
ALFWorld \cite{alfworld}, which involves moving between rooms and manipulating objects to complete household tasks;
and ScienceWorld \cite{scienceworld}, which requires conducting experiments in a virtual laboratory to achieve specified scientific goals. Appendix~\ref{app:prompt-templates} provides the prompt templates used for each benchmark.
Following the official implementation of TCOD \cite{tcod}, we use the 100-task WebShop evaluation set and the 1,308-task ScienceWorld evaluation set. The evaluation on ALFWorld covers the full official test set of 140 IID and 134 OOD tasks. Following Guided-OPD \cite{guided-opd}, we report task performance
using mean score and mean success rate (SR) for WebShop and ScienceWorld,
and SR alone for ALFWorld.
We also report the mean number of interaction
turns on each benchmark to measure deployment efficiency.

\paragraph{Training and evaluation setup.}

Most training and evaluation settings follow the official implementation of TCOD \cite{tcod}.
We train all methods using the Trinity-RFT
framework \cite{trinity-rft} with the
AdamW optimizer \cite{adamw}. We evaluate the final checkpoint using one rollout per task. Following TCOD and Guided-OPD, we use random seed 42 for all stochastic processes.
All experiments are conducted on eight NVIDIA A100 GPUs with 80\,GB of memory each.
Detailed training and evaluation settings are provided in
Appendix~\ref{app:full-configurations} (\cref{tab:shared-configurations}).

\paragraph{DASH-OPD settings.}
We set the intervention threshold ratio to $q_{\mathrm{on}}=0.1$
and the return threshold ratio to
$q_{\mathrm{off}}=q_{\mathrm{on}}/3$.
The minimum teacher segment length $\ell_{\min}$ is set to $2$,
and the stagnation threshold $K$ is set to $3$.
In \cref{eq:dash-standardization}, we lower-bound the denominator by $10^{-3}$ and clip $\bar d_t^{m_t}$ to $\left[-10,\,10\right]$. These hyperparameters are held fixed across all benchmarks
and model scales.

\begin{figure}[t]
    \centering
    \includegraphics[width=\linewidth]{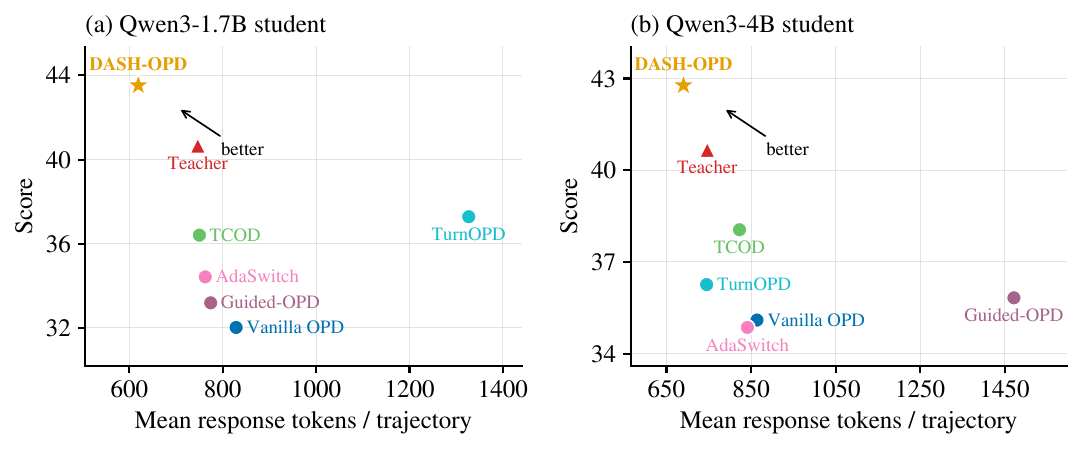}
    \caption{Performance--efficiency Pareto analysis on WebShop (mean score versus
    mean response tokens per task).
    Higher scores and fewer response tokens are better (upper left).
    At both model scales, DASH-OPD strictly Pareto-dominates all baselines and the zero-shot teacher.
    On ALFWorld and ScienceWorld, DASH-OPD is Pareto-optimal among baselines
    at 1.7B, and strictly Pareto-dominates all baselines (as well as the teacher on ALFWorld) at 4B
    (see \cref{fig:deployment-pareto-alfworld-app,fig:deployment-pareto-scienceworld-app} in Appendix~\ref{app:deployment-efficiency}).}
    \label{fig:deployment-pareto-webshop}
\end{figure}

\begin{figure}[!b]
    \centering
    \includegraphics[width=\linewidth]{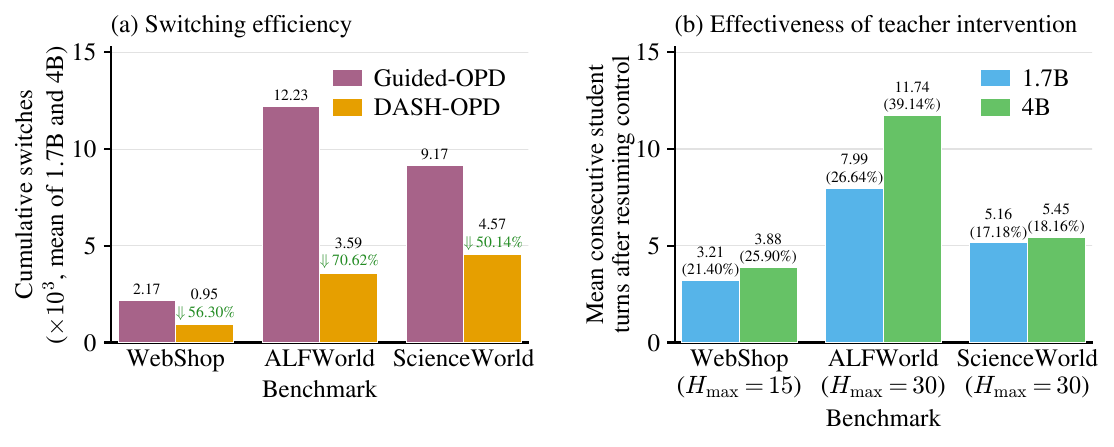}
    \caption{\textbf{(a)} Cumulative number of executor switches averaged over the 1.7B and 4B models. Fewer executor switches can reduce the overhead of control transfer. Across all benchmarks, DASH-OPD requires fewer switches than Guided-OPD.
    \textbf{(b)} Mean number of consecutive student turns after the teacher returns control during DASH-OPD training. Percentages indicate the fraction of the turn limit $H_{\max}$. Across all benchmarks and model scales, students sustain multi-turn execution after teacher intervention, indicating restored autonomy.}
    \label{fig:switch-efficiency-anatomy}
\end{figure}

\subsection{Main results}

\paragraph{Task performance.}
\Cref{tab:main-results-all-envs} reports task performance across the three
benchmarks and two model scales. DASH-OPD outperforms all five baselines in all 14 task-performance comparisons.
On WebShop, the 1.7B DASH-OPD student achieves a score of 43.53 and an SR of
13.00\%, exceeding the best baseline, TurnOPD, by 6.24 points and
1.00 percentage point, respectively.
The 4B DASH-OPD student achieves a score of 42.77 and an SR of 16.00\%,
with corresponding gains of 4.72 points and 1.50 percentage points
over the strongest baseline, TCOD.
The 1.7B and 4B DASH-OPD students also exceed the zero-shot teacher's score by
2.86 and 2.10 points.
On ALFWorld, DASH-OPD outperforms all baselines on both the IID and OOD
splits at both model scales. The 1.7B and 4B DASH-OPD students achieve overall SRs
of 32.85\% and 41.24\%, each exceeding the best baseline
at the same scale by 3.65 percentage points.
The 4B DASH-OPD student also exceeds the teacher's overall SR by
0.36 percentage points.
On ScienceWorld, the 1.7B DASH-OPD student achieves a score of 38.46 and an SR of
12.77\%, improving over the best baseline, AdaSwitch, by 2.61 points
and 1.76 percentage points, respectively.
The 4B DASH-OPD student achieves a score of 50.97 and an SR of 22.40\%, with
corresponding gains of 3.67 points and 2.06 percentage points over the strongest baseline for each metric.

\begin{wrapfigure}{r}{0.45\textwidth}
    \vspace{-1em}
    \centering
    \includegraphics[width=\linewidth]{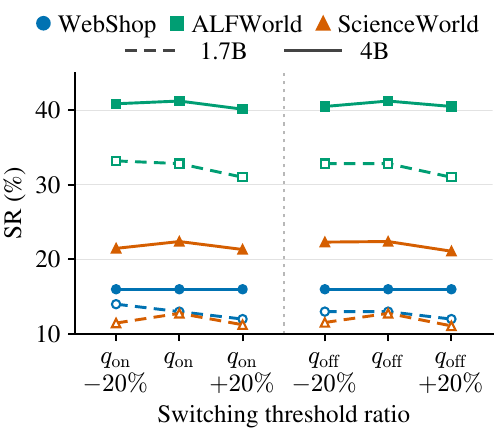}
    \caption{Sensitivity to switching threshold ratios.
    Each ratio is perturbed by $\pm20\%$ relative to its default while the other
    is held fixed. ALFWorld uses overall SR. Varying either hyperparameter changes SR by at most 1.82 percentage points, suggesting local robustness.}
    \label{fig:threshold-sensitivity}
\end{wrapfigure}

\paragraph{Deployment efficiency.}
\Cref{tab:main-results-all-envs} also reports deployment efficiency measured by
the mean number of interaction turns per task.
Among trained students, DASH-OPD requires the fewest turns in
nine of the ten efficiency comparisons: on WebShop and ALFWorld (IID, OOD, and overall)
at both model scales, and on ScienceWorld at 4B. 
On ScienceWorld at 1.7B, DASH-OPD ranks second with 23.60 turns, behind TCOD (22.80).

% Generated by latex/code/make_ablation_table.py from data/tab2_sources.csv.
% Raw metrics and provenance: data/tab2_metrics.csv.
\begin{table}[t]
    \centering
    \caption{Ablation results for DASH-OPD. Reporting conventions follow \cref{tab:main-results-all-envs}.
    \enquote{w/o teacher execution (OPD)}: use student-only rollouts without teacher intervention (i.e., vanilla OPD).
    \enquote{w/o student resumption}: prevent the student from resuming control after teacher intervention.
    \enquote{w/o signal accumulation}: use single-turn standardized signals instead of accumulated evidence and scale both switching threshold ratios by $0.4$.
    \enquote{w/o action mask}: compute teacher--student discrepancy over all response tokens instead of action tokens only.
    \enquote{w/o stagnation trigger}: disable teacher intervention triggered by stagnation in student actions or observations.
    \enquote{w/o minimum teacher span}: remove the minimum-length constraint on teacher-controlled segments.}
    \label{tab:ablation-results}
    \small
    \setlength{\tabcolsep}{2.1pt}
    \resizebox{\textwidth}{!}{%
    \begin{tabular}{@{}lccTcTcTcTccT@{}}
        \toprule
        \multirow{3}{*}{Method}
        & \multicolumn{3}{c}{WebShop}
        & \multicolumn{6}{c}{ALFWorld}
        & \multicolumn{3}{c}{ScienceWorld} \\
        \cmidrule(lr){2-4}\cmidrule(lr){5-10}\cmidrule(l){11-13}
        & \multicolumn{3}{c}{}
        & \multicolumn{2}{c}{IID} & \multicolumn{2}{c}{OOD}
        & \multicolumn{2}{c}{Overall} & \multicolumn{3}{c}{} \\
        \cmidrule(lr){5-6}\cmidrule(lr){7-8}\cmidrule(lr){9-10}
        & Score $\uparrow$ & SR $\uparrow$ & Turns $\downarrow$
        & SR $\uparrow$ & Turns $\downarrow$ & SR $\uparrow$ & Turns $\downarrow$
        & SR $\uparrow$ & Turns $\downarrow$
        & Score $\uparrow$ & SR $\uparrow$ & Turns $\downarrow$ \\
        \midrule
        \multicolumn{13}{@{}l}{Qwen3-1.7B student} \\
        % Train/eval (WebShop, ALFWorld, ScienceWorld): 110359/110584, 110273/110286, 111007/111007
        w/o teacher execution (OPD)
        & 32.03 & 10.00\% & 8.40
        & 23.57\% & 25.47 & 29.10\% & 25.01 & 26.28\% & 25.25
        & 27.23 & 7.65\% & 25.58 \\
        % Train/eval (WebShop, ALFWorld, ScienceWorld): 110732/110732, 110774/110774, 110734/110943
        w/o student resumption
        & 32.26 & \blue{\textbf{13.00\%}} & 8.39
        & 24.29\% & 25.84 & 31.34\% & 24.95 & 27.74\% & 25.40
        & 32.14 & 9.63\% & 25.42 \\
        % Train/eval (WebShop, ALFWorld, ScienceWorld): 110771/110779, 110772/110783, 110764/110764
        w/o signal accumulation
        & 35.90 & 10.00\% & 7.47
        & 27.86\% & 25.19 & 29.85\% & 25.18 & 28.83\% & 25.18
        & 34.01 & 10.32\% & 24.46 \\
        % Train/eval (WebShop, ALFWorld, ScienceWorld): 110645/110665, 110643/110662, 110685/110703
        w/o action mask
        & 38.47 & \underline{11.00\%} & 7.16
        & 27.14\% & 25.45 & 31.34\% & \underline{24.63} & 29.20\% & 25.05
        & \underline{37.75} & 11.39\% & 24.14 \\
        % Train/eval (WebShop, ALFWorld, ScienceWorld): 110598/110598, 110596/110602, 110640/110640
        w/o stagnation trigger
        & 39.83 & \underline{11.00\%} & 6.71
        & \blue{\textbf{32.86\%}} & \blue{\textbf{24.74}} & \underline{32.84\%} & \blue{\textbf{24.57}} & \blue{\textbf{32.85\%}} & \blue{\textbf{24.65}}
        & 37.04 & \underline{12.69\%} & \blue{\textbf{23.19}} \\
        % Train/eval (WebShop, ALFWorld, ScienceWorld): 110951/110969, 110687/110937, 110713/110713
        w/o minimum teacher span
        & \underline{41.15} & \underline{11.00\%} & \underline{6.52}
        & 26.43\% & 25.42 & \blue{\textbf{34.33\%}} & 24.81 & \underline{30.29\%} & 25.12
        & 37.59 & 12.16\% & 23.92 \\
        % Train/eval (WebShop, ALFWorld, ScienceWorld): 110399/110399, 110333/110333, 110343/110465
        \textbf{DASH-OPD}
        & \blue{\textbf{43.53}} & \blue{\textbf{13.00\%}} & \blue{\textbf{6.25}}
        & \underline{31.43\%} & \underline{24.98} & \blue{\textbf{34.33\%}} & 24.69 & \blue{\textbf{32.85\%}} & \underline{24.84}
        & \blue{\textbf{38.46}} & \blue{\textbf{12.77\%}} & \underline{23.60} \\
        \midrule
        \multicolumn{13}{@{}l}{Qwen3-4B student} \\
        % Train/eval (WebShop, ALFWorld, ScienceWorld): 110365/110365, 110290/110290, 111004/111006
        w/o teacher execution (OPD)
        & 35.10 & 12.00\% & 8.44
        & 34.29\% & 24.38 & 38.81\% & 24.40 & 36.50\% & 24.39
        & 43.71 & 16.36\% & 23.19 \\
        % Train/eval (WebShop, ALFWorld, ScienceWorld): 110773/110773, 110775/110775, 110731/110741
        w/o student resumption
        & 35.74 & 13.00\% & 8.35
        & 37.86\% & \underline{23.35} & 34.33\% & 24.54 & 36.13\% & 23.93
        & 46.92 & 20.11\% & 21.66 \\
        % Train/eval (WebShop, ALFWorld, ScienceWorld): 110766/110766, 110765/110765, 110797/110806
        w/o signal accumulation
        & 34.13 & 11.00\% & 8.87
        & 33.57\% & 24.62 & 36.57\% & 24.05 & 35.04\% & 24.34
        & 45.37 & 19.42\% & 21.75 \\
        % Train/eval (WebShop, ALFWorld, ScienceWorld): 110646/110666, 110644/110940, 110670/110670
        w/o action mask
        & 37.85 & \underline{15.00\%} & 7.87
        & 37.14\% & 24.37 & \blue{\textbf{41.79\%}} & \underline{23.40} & 39.42\% & 23.89
        & \underline{49.06} & \underline{21.71\%} & 21.02 \\
        % Train/eval (WebShop, ALFWorld, ScienceWorld): 110601/110601, 110599/110608, 110600/110612
        w/o stagnation trigger
        & \underline{40.39} & \blue{\textbf{16.00\%}} & 7.63
        & \underline{40.71\%} & 23.74 & 37.31\% & 24.60 & 39.05\% & 24.16
        & 46.14 & \blue{\textbf{22.40\%}} & \blue{\textbf{20.61}} \\
        % Train/eval (WebShop, ALFWorld, ScienceWorld): 110690/110690, 110688/110701, 110710/110946
        w/o minimum teacher span
        & 39.58 & \blue{\textbf{16.00\%}} & \underline{7.57}
        & \blue{\textbf{41.43\%}} & 23.42 & 38.06\% & \blue{\textbf{23.27}} & \underline{39.78\%} & \underline{23.35}
        & 47.70 & 21.10\% & 21.59 \\
        % Train/eval (WebShop, ALFWorld, ScienceWorld): 110401/110456, 110310/110310, 110344/110467
        \textbf{DASH-OPD}
        & \blue{\textbf{42.77}} & \blue{\textbf{16.00\%}} & \blue{\textbf{7.15}}
        & \blue{\textbf{41.43\%}} & \blue{\textbf{22.86}} & \underline{41.04\%} & 23.46 & \blue{\textbf{41.24\%}} & \blue{\textbf{23.15}}
        & \blue{\textbf{50.97}} & \blue{\textbf{22.40\%}} & \underline{20.62} \\
        \bottomrule
    \end{tabular}%
    }
\end{table}

\paragraph{Performance--efficiency Pareto analysis.}
\Cref{fig:deployment-pareto-webshop} compares mean score and mean
response tokens per task on WebShop. At both model scales, DASH-OPD strictly Pareto-dominates all baselines and the zero-shot teacher, achieving higher scores
with fewer response tokens. The 1.7B and 4B DASH-OPD students achieve scores of
43.53 and 42.77 with 619.24 and 690.24 response tokens,
respectively. Compared with the teacher, they improve
scores by 2.86 and 2.10 points while reducing response tokens by 127.37 and 56.37. The corresponding comparisons on ALFWorld and ScienceWorld are provided in
Appendix~\ref{app:deployment-efficiency}
(\cref{fig:deployment-pareto-alfworld-app,fig:deployment-pareto-scienceworld-app}).
On both benchmarks, the 1.7B DASH-OPD students are Pareto-optimal among baselines, while the 4B students strictly
Pareto-dominate all baselines. On ALFWorld, the 4B student also Pareto-dominates the teacher.
Appendix~\ref{app:deployment-tokens} reports detailed token counts for all methods, including zero-shot students.

\paragraph{Number of executor switches.}
\Cref{fig:switch-efficiency-anatomy}(a) compares the cumulative number of
executor switches for DASH-OPD and Guided-OPD (averaged over the 1.7B and 4B models). Fewer executor switches can reduce the overhead of transferring control between the student and teacher. DASH-OPD consistently requires fewer switches than Guided-OPD across all three benchmarks. This advantage holds at both model scales. On WebShop, DASH-OPD reduces the total switches from 2,186 to 901 at 1.7B and from 2,155 to 996 at 4B,
corresponding to reductions of 58.78\% and 53.78\%, respectively.
On ALFWorld, DASH-OPD records 9,172 fewer switches (73.88\%) at 1.7B
and 8,105 fewer (67.25\%) at 4B. On ScienceWorld, the corresponding
reductions are 4,625 (49.02\%) and 4,574 (51.32\%).

DASH-OPD's lower switch count stems from its hysteretic switching mechanism, which suppresses frequent alternation between the teacher and student.
Appendix~\ref{app:hysteretic-switching} provides further evidence of the hysteresis.

\paragraph{Effectiveness of teacher intervention.}
\Cref{fig:switch-efficiency-anatomy}(b) reports the mean number of consecutive student turns
after the teacher returns control during DASH-OPD training.
The turn limit per trajectory ($H_{\max}$) is $15$ for WebShop and $30$ for ALFWorld and ScienceWorld.
On WebShop, the 1.7B and 4B students remain in control for an average of 3.21 and 3.88 turns, respectively, corresponding to 21.40\% and 25.90\% of the turn limit.
The corresponding spans are 7.99 and 11.74 turns on ALFWorld
(26.64\% and 39.14\% of its turn limit) and 5.16 and 5.45 turns on
ScienceWorld (17.18\% and 18.16\%).
Across all benchmarks, students at both scales execute multiple consecutive turns after regaining control, suggesting that teacher intervention helps restore sustained student autonomy.
The 4B students maintain control longer than the 1.7B students, likely because of their greater capacity.
Appendix~\ref{app:student-survival-after-resuming-control} complements this analysis with student survival rates after the teacher returns control.

\subsection{Ablation study}
\label{sec:ablation-study}

\paragraph{Core components.}
We ablate three core components of DASH-OPD: teacher intervention during training rollouts, student resumption of control, and cross-turn signal accumulation. The results are reported in the \enquote{w/o teacher execution (OPD)}, \enquote{w/o student resumption}, and \enquote{w/o signal accumulation} rows in \cref{tab:ablation-results}, respectively.
In the \enquote{w/o signal accumulation} ablation, single-turn signals
are typically smaller in magnitude than accumulated evidence, making
the original switching thresholds harder to reach. We therefore scale both $q_{\mathrm{on}}$ and $q_{\mathrm{off}}$ by $0.4$ (selected using the same tuning budget as for full DASH-OPD).
Removing any of these core components substantially reduces task performance
at both model scales: scores drop by 7.03--11.50 points on WebShop and by 4.05--11.23 points on ScienceWorld, while overall SR drops by 4.02--6.57 percentage points on ALFWorld.
Appendix~\ref{app:training-dynamics} shows decreasing trends in teacher--student discrepancy and distillation losses over the course of training, supporting the effectiveness of DASH-OPD training.

\paragraph{Additional components.}
We also ablate three additional components of DASH-OPD: action masking in
teacher--student discrepancy computation, stagnation-triggered teacher intervention, and the minimum-length constraint on teacher-controlled segments.
The results are reported in the \enquote{w/o action mask},
\enquote{w/o stagnation trigger}, and \enquote{w/o minimum teacher span}
rows in \cref{tab:ablation-results}, respectively.
Removing these components generally incurs smaller performance losses
than the core ablations: scores drop by
2.38--5.06 points on WebShop and by 0.71--4.83 points on ScienceWorld,
while overall SR drops by 0.00--3.65 percentage points on ALFWorld.

\paragraph{Hyperparameter sensitivity.}
We assess sensitivity to the switching threshold ratios by perturbing $q_{\mathrm{on}}$ and $q_{\mathrm{off}}$ individually by $\pm20\%$ relative to their default values, while holding the other ratio fixed. As shown in \cref{fig:threshold-sensitivity},
across all benchmarks and model scales, SR differs
from the default setting by at most 1.82 percentage points.
On WebShop, the 4B model maintains 16.00\% SR under every perturbation.
These results suggest robustness to local changes in both hyperparameters.

\section{Conclusion}

We propose DASH-OPD, the first agentic OPD method to perform adaptive, bidirectional executor switching. It uses directional teacher--student discrepancy to build drift and recovery evidence, then compares the evidence with thresholds to guide switching between the teacher and student. This hysteretic switching mechanism prevents rapid alternation between the two models. Across three benchmarks and two model scales, DASH-OPD outperforms five baselines in all task-performance comparisons and requires the fewest interaction turns in nine of ten efficiency comparisons.

% \newpage % 需要注释

\bibliographystyle{refs/IEEEtran}
\bibliography{refs/IEEEabrv, refs/IEEEexample}

%%%%%%%%%%%%%%%%%%%%%%%%%%%%%%%%%%%%%%%%%%%%%%%%%%%%%%%%%%%%

\clearpage
\appendix
\section*{Appendix} % 需要注释
\section{DASH-OPD algorithmic details}
\label{app:algorithm}

\Cref{alg:dash-opd} presents the pseudocode for DASH-OPD. For improved training effectiveness, DASH-OPD does not follow TCOD and Guided-OPD in gradually reverting to vanilla OPD.
Given the teacher--student discrepancy, DASH-OPD updates its statistics
and switching state in $O(1)$ time per turn and uses $O(1)$ memory per trajectory.

\begin{algorithm}[h]
\caption{DASH-OPD}
\label{alg:dash-opd}
\begin{algorithmic}[1]
\Require Student $\pi_\theta$, rollout student $\pi_{\bar\theta}$, teacher $\pi_T$; environment $\mathcal E$ and initial history $h_1$; training step $s$ and total steps $S_{\max}$; turn limit $H_{\max}$; $q_{\mathrm{on}},q_{\mathrm{off}},\ell_{\min},K$
\State Set $\tau_{\mathrm{on}},\tau_{\mathrm{off}}$ by \cref{eq:dash-thresholds}; set $D=R=\ell=0$ and initialize running statistics $\mathcal R^S,\mathcal R^T$
\State Sample $m_1=T$ with probability $p_T(s)$ (\cref{eq:teacher-start}); otherwise set $m_1=S$
\For{$t=1,\ldots,H_{\max}$}
    \State Generate response $y_t$ under executor $m_t$ given $h_t$ (\cref{eq:dash-generation})
    \State Compute action-token discrepancy $d_t^{m_t}$ (\cref{eq:student-ratio,eq:teacher-ratio})
    \State Compute $\bar d_t^{m_t}$ using prior $\mathcal R^{m_t}$ (\cref{eq:dash-standardization}); update $\mathcal R^{m_t}$ with $d_t^{m_t}$
    \State Execute the action in $y_t$ in $\mathcal E$ and update $h_{t+1}$; \textbf{break} if the episode ends
    \State $m_{t+1}\gets m_t$
    \If{$m_t=S$}
        \State $D\gets\max(D+\bar d_t^S,0)$; compute stagnation flag $g_t$ (\cref{eq:support-switching})
        \State \textbf{if} $D>\tau_{\mathrm{on}}\lor g_t$ \textbf{then} $m_{t+1}\gets T$; $D,\ell\gets0$
    \Else
        \State $R\gets\max(R-\bar d_t^T,0)$; $\ell\gets\ell+1$
        \State \textbf{if} $R>\tau_{\mathrm{off}}\land\ell\ge\ell_{\min}$ \textbf{then} $m_{t+1}\gets S$; $R,\ell\gets0$
    \EndIf
\EndFor
\State Update $\pi_\theta$ on the collected responses: sampled reverse KL on student turns and sampled forward KL on teacher turns (\cref{eq:dash-objective})
\end{algorithmic}
\end{algorithm}

\section{Detailed training objectives for Guided-OPD and DASH-OPD}
\label{app:token-losses}

Following Guided-OPD, DASH-OPD uses the same
role-specific token losses and clipping settings. Let
$z_{t,i}^{j}=\log\pi_j(y_{t,i}\mid h_t,y_{t,<i})$ for
$j\in\{\theta,\bar\theta,T\}$. On a student turn, we optimize a PPO-clipped
objective \cite{ppo} using the advantage $a_{t,i}=z_{t,i}^{T}-z_{t,i}^{\bar\theta}$. Define the policy
ratio $r_{t,i}=\exp\bigl(\operatorname{clip}(z_{t,i}^{\theta}-z_{t,i}^{\bar\theta},-20,20)\bigr)$,
its clipped value $\tilde r_{t,i}=\operatorname{clip}(r_{t,i},1-\epsilon,1+\epsilon)$,
and $u_{t,i}=\max\{-a_{t,i}r_{t,i},-a_{t,i}\tilde r_{t,i}\}$.
The student-turn token loss in \cref{eq:dash-objective} can be expressed as the following dual-clipped PPO loss \cite{ye2020mastering}:
\begin{equation}
\mathcal L_{t,i}^{S}=
\begin{cases}
u_{t,i}, & a_{t,i}\ge 0,\\
\min\{u_{t,i},-c\,a_{t,i}\}, & a_{t,i}<0.
\end{cases}
\label{eq:student-turn-token-loss}
\end{equation}
On a teacher turn, the teacher's log-probability is independent
of $\theta$. Minimizing the sampled forward KL is therefore equivalent to
minimizing the following token-level negative log-likelihood (NLL):
\begin{equation}
\mathcal L_{t,i}^{T}=-z_{t,i}^{\theta}
=-\log\pi_\theta(y_{t,i}\mid h_t,y_{t,<i}).
\label{eq:teacher-turn-token-loss}
\end{equation}
Both methods use $\epsilon=0.2$, a dual-clip coefficient $c=3$, and a KL
coefficient of $1$. Sequence masking and policy-gradient fallback are
disabled (see \cref{tab:baseline-configurations}). The inner clip in
$r_{t,i}$ limits the log ratio for numerical stability.
\Cref{eq:dash-objective} averages these losses over valid response tokens.

\section{Rationale for the directionality of teacher--student discrepancy}
\label{app:directional-discrepancy}

In \cref{eq:student-ratio,eq:teacher-ratio}, we place the executing policy in
the numerator of the log-likelihood ratio. Consequently, the
student-to-teacher ratio $d_t^S$ is a sample-based analogue of
$D_{\mathrm{KL}}(\pi_{\bar\theta}\Vert\pi_T)$, whereas the
teacher-to-student ratio $d_t^T$ is a sample-based analogue of
$D_{\mathrm{KL}}(\pi_T\Vert\pi_{\bar\theta})$. Both ratios therefore share
a common interpretation: larger values indicate greater discrepancy with the
non-executing model. Their directions match the sampled reverse- and forward-KL objectives on student- and teacher-generated responses, respectively, and both ratios are directly computed from the log-probabilities collected during rollout scoring without further model forward passes.

\section{Training dynamics}
\label{app:training-dynamics}

\begin{figure}[h]
    \centering
    \includegraphics[width=\linewidth]{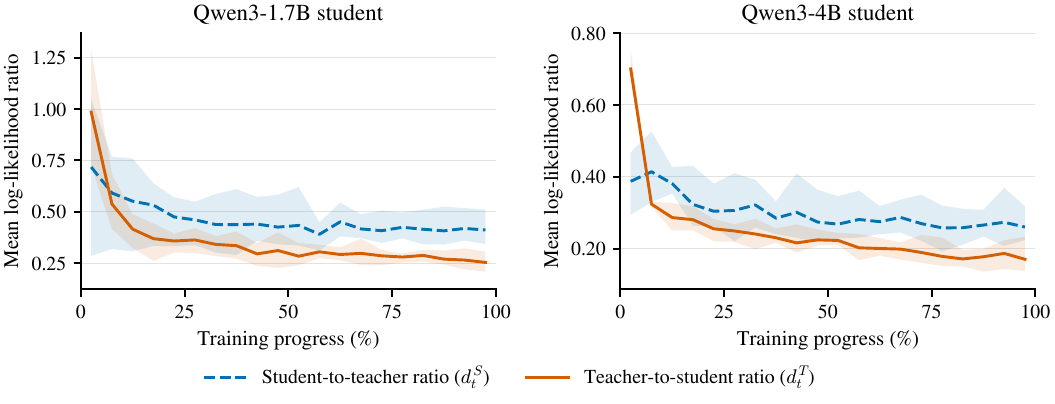}
    \caption{Directional teacher--student discrepancy during DASH-OPD training. These discrepancies are measured by the student-to-teacher and teacher-to-student mean log-likelihood ratios in \cref{eq:student-ratio,eq:teacher-ratio}. The results are aggregated across all three benchmarks.
    \enquote{Training progress} is measured relative to the planned 150 training steps on WebShop and 250
    steps on ALFWorld and ScienceWorld, then divided into 20 equal intervals (5\% each).
    For each interval and model scale, we take the median batch estimate within each benchmark.
    The line shows the mean across benchmarks, and the shading spans the minimum and maximum. The decreasing discrepancies in both directions suggest the effectiveness of the distillation.}
    \label{fig:appendix-directional-discrepancy}
\end{figure}

As shown in \cref{fig:appendix-directional-discrepancy}, the aggregate
directional teacher--student discrepancy decreases over training at both model
scales. From the first to the last 5\% training-progress interval,
the benchmark-mean student-to-teacher ratio falls from 0.72 to
0.41 (42.65\%) for 1.7B and from 0.39 to 0.26 (32.80\%) for 4B. The
teacher-to-student ratio falls from 0.99 to 0.25 (74.20\%) at 1.7B and from
0.70 to 0.17 (75.60\%) at 4B. These reductions provide evidence for the
distillation process.

\begin{figure}[h]
    \centering
    \includegraphics[width=\linewidth]{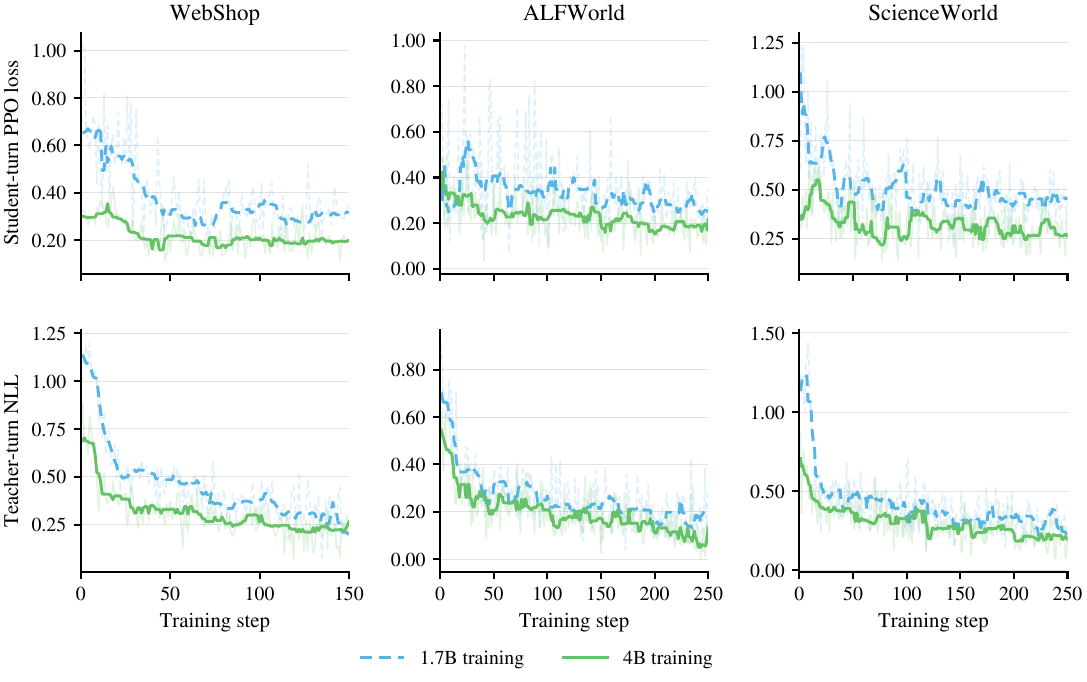}
    \caption{Distillation losses during DASH-OPD training. The top and bottom rows show the logged, token-averaged values corresponding to the student-turn PPO loss in \cref{eq:student-turn-token-loss} and the teacher-turn negative log-likelihood (NLL) in \cref{eq:teacher-turn-token-loss}, respectively. Thin lines show logged
    step values, and thick lines show centered nine-step running medians. The decreasing losses indicate the effectiveness of the training process.}
    \label{fig:appendix-training-loss-components}
\end{figure}

\Cref{fig:appendix-training-loss-components} traces the two distillation losses detailed in \cref{app:token-losses}. Their decreasing trends suggest that the model is being effectively optimized throughout training.

\section{Comparison with TCOD variants}
\label{app:tcod-variants}

\Cref{tab:tcod-variants} reports TCOD-B2F and TCOD-F2B separately. DASH-OPD
exceeds both variants in all 14 task-performance comparisons across the three
benchmarks and two model scales. Against the stronger variant on each metric,
DASH-OPD's WebShop score gains are 6.03 and 3.99 points at 1.7B and 4B, and its
ScienceWorld score gains are 7.48 and 3.29 points. The corresponding gains
in ALFWorld overall SR are 2.19 and 4.01 percentage points. In addition, DASH-OPD uses the fewest turns in nine of ten efficiency comparisons (second fewest on ScienceWorld at 1.7B).

% Sources follow the training/evaluation pairs in tabs/id.tex.
% Values are rounded from full-precision metrics (half up, two decimals),
% removing binary floating-point noise at 12 decimals before final rounding.
% Rankings and all prose comparisons use the displayed two-decimal values.
\begin{table}[!b]
    \centering
    \caption{Task performance and mean interaction turns for TCOD-B2F,
    TCOD-F2B, and DASH-OPD across three benchmarks and two student scales.
    B2F and F2B mean backward-to-forward and forward-to-backward. ALFWorld
    \enquote{Overall} combines the IID and OOD test splits. Higher scores and
    SRs are better; fewer turns are better.
    Within each model scale, \blue{\textbf{blue bold}} and
    \underline{underline} mark the best and second-best results,
    respectively (ties share the same style). DASH-OPD leads all 14 task-performance comparisons
    and requires the fewest turns in 9 of 10 efficiency comparisons (ranking second in the remaining one).}
    \label{tab:tcod-variants}
    \small
    \setlength{\tabcolsep}{2.1pt}
    \resizebox{\textwidth}{!}{%
    \begin{tabular}{@{}lccTcTcTcTccT@{}}
        \toprule
        \multirow{3}{*}{Method}
        & \multicolumn{3}{c}{WebShop}
        & \multicolumn{6}{c}{ALFWorld}
        & \multicolumn{3}{c}{ScienceWorld} \\
        \cmidrule(lr){2-4}\cmidrule(lr){5-10}\cmidrule(l){11-13}
        & \multicolumn{3}{c}{}
        & \multicolumn{2}{c}{IID} & \multicolumn{2}{c}{OOD}
        & \multicolumn{2}{c}{Overall} & \multicolumn{3}{c}{} \\
        \cmidrule(lr){5-6}\cmidrule(lr){7-8}\cmidrule(lr){9-10}
        & Score $\uparrow$ & SR $\uparrow$ & Turns $\downarrow$
        & SR $\uparrow$ & Turns $\downarrow$ & SR $\uparrow$ & Turns $\downarrow$
        & SR $\uparrow$ & Turns $\downarrow$
        & Score $\uparrow$ & SR $\uparrow$ & Turns $\downarrow$ \\
        \midrule
        \multicolumn{13}{@{}l}{Qwen3-1.7B student} \\
        % checkpoints/WEBSHOP/webshop_tcod_b2f_110364/benchmark_results/bench_results_job_110364.json
        % checkpoints/ALFWORLD/alfworld_tcod_b2f_110282/monitor/local_metrics/benchmark.jsonl:396
        % checkpoints/SCIENCEWORLD/scienceworld_tcod_b2f_110341/benchmark_results/bench_results_job_110341.json
        TCOD-B2F
        & 35.33 & \underline{11.00\%} & 7.80
        & 22.86\% & 25.64 & 31.34\% & 24.99 & 27.01\% & 25.32
        & 29.91 & 3.44\% & \blue{\textbf{20.15}} \\
        % checkpoints/WEBSHOP/webshop_tcod_f2b_110361/benchmark_results/bench_results_job_110375.json
        % checkpoints/ALFWORLD/alfworld_tcod_f2b_110450/benchmark_results/bench_results_job_110450.json
        % checkpoints/SCIENCEWORLD/scienceworld_tcod_f2b_110337/benchmark_results/bench_results_job_110337.json
        TCOD-F2B
        & \underline{37.50} & \underline{11.00\%} & \underline{7.56}
        & \underline{28.57\%} & \underline{25.20} & \underline{32.84\%} & \underline{24.98} & \underline{30.66\%} & \underline{25.09}
        & \underline{30.98} & \underline{8.41\%} & 25.45 \\
        % checkpoints/WEBSHOP/webshop_dash_opd_110399/benchmark_results/bench_results_job_110399.json
        % checkpoints/ALFWORLD/alfworld_dash_opd_110333/benchmark_results/bench_results_job_110333.json
        % checkpoints/SCIENCEWORLD/scienceworld_dash_opd_110343/benchmark_results/bench_results_job_110465.json
        \textbf{DASH-OPD}
        & \blue{\textbf{43.53}} & \blue{\textbf{13.00\%}} & \blue{\textbf{6.25}}
        & \blue{\textbf{31.43\%}} & \blue{\textbf{24.98}} & \blue{\textbf{34.33\%}} & \blue{\textbf{24.69}} & \blue{\textbf{32.85\%}} & \blue{\textbf{24.84}}
        & \blue{\textbf{38.46}} & \blue{\textbf{12.77\%}} & \underline{23.60} \\
        \midrule
        \multicolumn{13}{@{}l}{Qwen3-4B student} \\
        % checkpoints/WEBSHOP/webshop_tcod_b2f_110366/benchmark_results/bench_results_job_110366.json
        % checkpoints/ALFWORLD/alfworld_tcod_b2f_110302/benchmark_results/bench_results_job_110302.json
        % checkpoints/SCIENCEWORLD/scienceworld_tcod_b2f_110347/benchmark_results/bench_results_job_110347.json
        TCOD-B2F
        & 37.32 & 14.00\% & 8.34
        & \underline{37.86\%} & \underline{23.59} & 32.84\% & 24.29 & 35.40\% & 23.93
        & \underline{47.68} & 19.27\% & 22.13 \\
        % checkpoints/WEBSHOP/webshop_tcod_f2b_110404/benchmark_results/bench_results_job_110404.json
        % checkpoints/ALFWORLD/alfworld_tcod_f2b_110451/benchmark_results/bench_results_job_110454.json
        % checkpoints/SCIENCEWORLD/scienceworld_tcod_f2b_110452/benchmark_results/bench_results_job_110452.json
        TCOD-F2B
        & \underline{38.78} & \underline{15.00\%} & \underline{7.70}
        & \underline{37.86\%} & 23.70 & \underline{36.57\%} & \underline{24.10} & \underline{37.23\%} & \underline{23.90}
        & 46.92 & \underline{20.11\%} & \underline{21.82} \\
        % checkpoints/WEBSHOP/webshop_dash_opd_110401/benchmark_results/bench_results_job_110456.json
        % checkpoints/ALFWORLD/alfworld_dash_opd_110310/benchmark_results/bench_results_job_110310.json
        % checkpoints/SCIENCEWORLD/scienceworld_dash_opd_110344/benchmark_results/bench_results_job_110467.json
        \textbf{DASH-OPD}
        & \blue{\textbf{42.77}} & \blue{\textbf{16.00\%}} & \blue{\textbf{7.15}}
        & \blue{\textbf{41.43\%}} & \blue{\textbf{22.86}} & \blue{\textbf{41.04\%}} & \blue{\textbf{23.46}} & \blue{\textbf{41.24\%}} & \blue{\textbf{23.15}}
        & \blue{\textbf{50.97}} & \blue{\textbf{22.40\%}} & \blue{\textbf{20.62}} \\
        \bottomrule
    \end{tabular}%
    }
\end{table}

\section{Performance--efficiency Pareto analysis on ALFWorld and ScienceWorld}
\label{app:deployment-efficiency}

\begin{figure}[t]
    \centering
    \includegraphics[width=\linewidth]{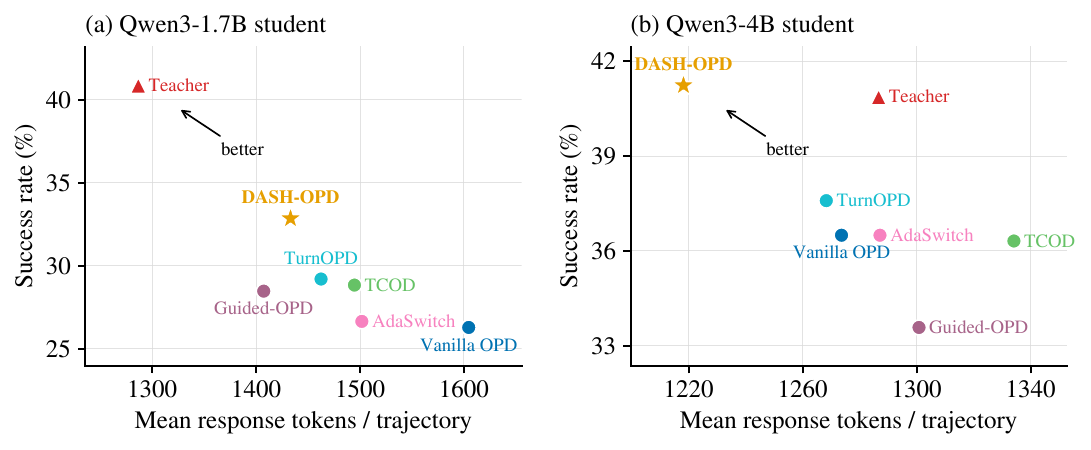}
    \caption{Performance--efficiency Pareto analysis on ALFWorld (overall SR versus
    mean response tokens per task). Higher SR and fewer response tokens are
    better (upper left). The 1.7B DASH-OPD student is Pareto-optimal among
    baselines, and the 4B student strictly Pareto-dominates all baselines and the
    zero-shot teacher.}
    \label{fig:deployment-pareto-alfworld-app}
\end{figure}

\Cref{fig:deployment-pareto-alfworld-app} compares overall SR and mean response
tokens per task on the full 274-task ALFWorld test set. The 1.7B DASH-OPD
student is Pareto-optimal among baselines, achieving 32.85\% overall SR with
1,433.13 response tokens per task.
Compared with TurnOPD, the strongest baseline by SR, it gains 3.65
percentage points and uses 29.37 fewer tokens. The 4B student strictly
Pareto-dominates all baselines and the zero-shot teacher, achieving 41.24\% SR
with 1,218.21 tokens. Relative to TurnOPD, it gains 3.65 percentage points and
uses 50.11 fewer tokens. Relative to the teacher, it gains 0.36 percentage
points and uses 68.46 fewer tokens.

\begin{figure}[t]
    \centering
    \includegraphics[width=\linewidth]{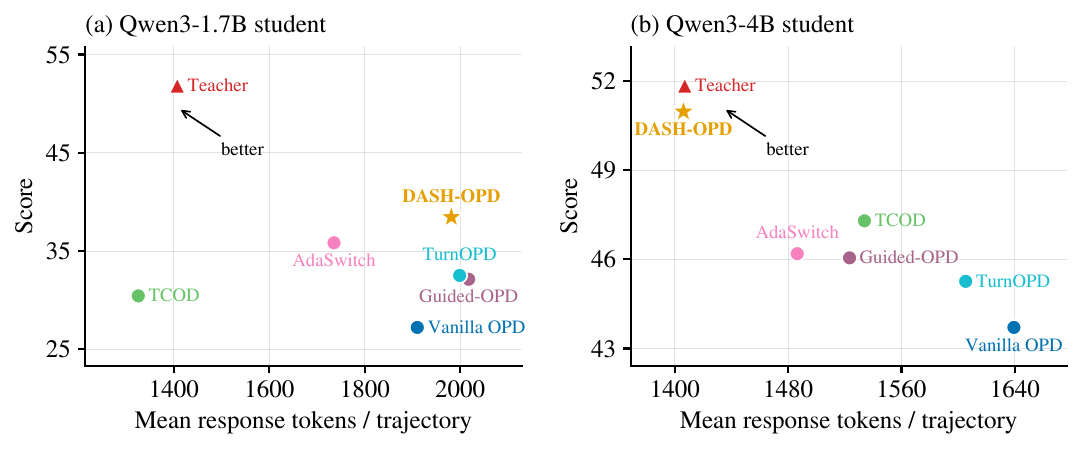}
    \caption{Performance--efficiency Pareto analysis on ScienceWorld (mean score versus
    mean response tokens per task). Higher scores and fewer response tokens are
    better (upper left). The 1.7B DASH-OPD student is Pareto-optimal among
    baselines, and the 4B student strictly Pareto-dominates all baselines.}
    \label{fig:deployment-pareto-scienceworld-app}
\end{figure}

% Plotted methods: sources recorded in data/fig_deployment_pareto_<environment>.csv.
% Values rounded half up to two decimals from the original benchmark metrics;
% TCOD averages the unrounded B2F and F2B metrics before rounding.
% Rankings and prose comparisons use the displayed values.
% Zero-shot webshop 1.7B: checkpoints/WEBSHOP/webshop_opd_model_bench_110357/benchmark_results/bench_results_job_110357.json
% Zero-shot webshop 4B: checkpoints/WEBSHOP/webshop_opd_model_bench_110358/benchmark_results/bench_results_job_110358.json
% Zero-shot alfworld 1.7B: checkpoints/ALFWORLD/alfworld_opd_model_bench_110283/monitor/local_metrics/benchmark.jsonl:396
% Zero-shot alfworld 4B: checkpoints/ALFWORLD/alfworld_opd_model_bench_110296/benchmark_results/bench_results_job_110296.json
% Zero-shot scienceworld 1.7B: checkpoints/SCIENCEWORLD/scienceworld_opd_model_bench_110349/benchmark_results/bench_results_job_110349.json
% Zero-shot scienceworld 4B: checkpoints/SCIENCEWORLD/scienceworld_opd_model_bench_110350/benchmark_results/bench_results_job_110350.json
\begin{table}[t]
    \centering
    \caption{Mean response tokens per task across three benchmarks and two
    student scales. ALFWorld includes all 274 IID and OOD test tasks.
    TCOD entries average its B2F and F2B variants. Fewer tokens are better.
    Zero-shot teacher and student results are shown for reference and excluded
    from rankings.
    Within each model scale, \blue{\textbf{blue bold}} and
    \underline{underline} mark the best and second-best results,
    respectively (ties share the same style).}
    \label{tab:deployment-tokens}
    \normalsize
    \setlength{\tabcolsep}{6pt}
    \begin{tabularx}{\textwidth}{@{}>{\raggedright\arraybackslash}X*{3}{>{\raggedleft\arraybackslash}X}@{}}
        \toprule
        Method & WebShop & ALFWorld & ScienceWorld \\
        & Tokens $\downarrow$ & Tokens $\downarrow$ & Tokens $\downarrow$ \\
        \midrule
        \multicolumn{4}{@{}l}{Qwen3-30B-A3B teacher} \\
        \gray{Zero-shot} & \gray{746.61} & \gray{1,286.67} & \gray{1,407.07} \\
        \midrule
        \multicolumn{4}{@{}l}{Qwen3-1.7B student} \\
        \gray{Zero-shot} & \gray{456.36} & \gray{510.76} & \gray{186.65} \\
        Vanilla OPD & 828.66 & 1,604.59 & 1,910.01 \\
        AdaSwitch & 762.52 & 1,501.75 & \underline{1,735.51} \\
        TCOD & \underline{750.11} & 1,494.63 & \blue{\textbf{1,325.41}} \\
        Guided-OPD & 774.27 & \blue{\textbf{1,407.35}} & 2,017.77 \\
        TurnOPD & 1,326.80 & 1,462.50 & 1,998.78 \\
        \textbf{DASH-OPD} & \blue{\textbf{619.24}} & \underline{1,433.13} & 1,981.40 \\
        \midrule
        \multicolumn{4}{@{}l}{Qwen3-4B student} \\
        \gray{Zero-shot} & \gray{1,113.30} & \gray{1,301.18} & \gray{1,428.46} \\
        Vanilla OPD & 864.06 & 1,273.66 & 1,639.35 \\
        AdaSwitch & 841.39 & 1,287.15 & \underline{1,486.43} \\
        TCOD & 822.71 & 1,334.16 & 1,533.88 \\
        Guided-OPD & 1,471.92 & 1,300.85 & 1,523.34 \\
        TurnOPD & \underline{744.95} & \underline{1,268.32} & 1,605.32 \\
        \textbf{DASH-OPD} & \blue{\textbf{690.24}} & \blue{\textbf{1,218.21}} & \blue{\textbf{1,406.16}} \\
        \bottomrule
    \end{tabularx}
\end{table}

\begin{figure}[!b]
    \centering
    \includegraphics[width=\linewidth]{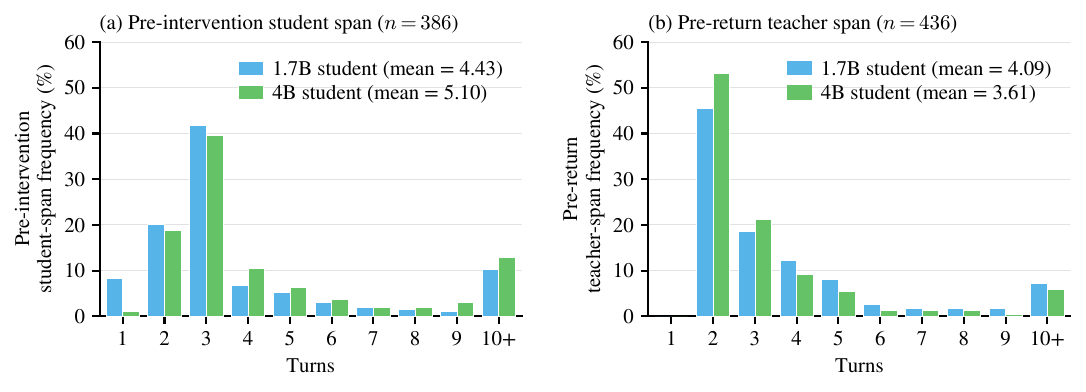}
    \caption{Number of consecutive student or teacher turns before an
    executor switch. Rollouts are sampled every 50 training steps from the
    six DASH-OPD training runs in \cref{tab:main-results-all-envs}.
    \textbf{(a)} Student turns before teacher intervention.
    \textbf{(b)} Teacher turns before control returns to the student.
    The most common spans are three student turns before teacher
    intervention and two teacher turns before control returns to the
    student, consistent with DASH-OPD's hysteretic switching mechanism.}
    \label{fig:preceding-run-lengths}
\end{figure}

\Cref{fig:deployment-pareto-scienceworld-app} compares mean score and mean
response tokens per task on the 1,308-task ScienceWorld test set. The 1.7B
DASH-OPD student is Pareto-optimal among baselines, scoring 38.46 with 1,981.40
tokens per task. Relative to AdaSwitch, the strongest baseline by score, it
gains 2.61 points at a cost of 245.89 more tokens. The 4B student strictly
Pareto-dominates all baselines, scoring 50.97 with 1,406.16 tokens. Relative
to TCOD, the strongest baseline by score, it gains 3.67 points while using
127.72 fewer tokens.

\section{Detailed response-token results}
\label{app:deployment-tokens}

\Cref{tab:deployment-tokens} gives the token counts underlying
\cref{fig:deployment-pareto-webshop,fig:deployment-pareto-alfworld-app,fig:deployment-pareto-scienceworld-app} and includes the zero-shot students in \cref{tab:main-results-all-envs} for reference.
For each task, we sum response tokens across interaction turns, then average
the task totals over the test set.

Among trained students, DASH-OPD uses the fewest tokens in four of six
response-token comparisons (second fewest on ALFWorld at 1.7B): WebShop at both scales, plus ALFWorld and
ScienceWorld at 4B. At 1.7B, it uses 25.78 more tokens than Guided-OPD on
ALFWorld and 655.99 more than TCOD on ScienceWorld, while gaining 4.38
percentage points in overall SR and 8.01 points in mean score, respectively.

\section{Hysteretic switching}
\label{app:hysteretic-switching}

\begin{figure}[t]
    \centering
    \includegraphics[width=\linewidth]{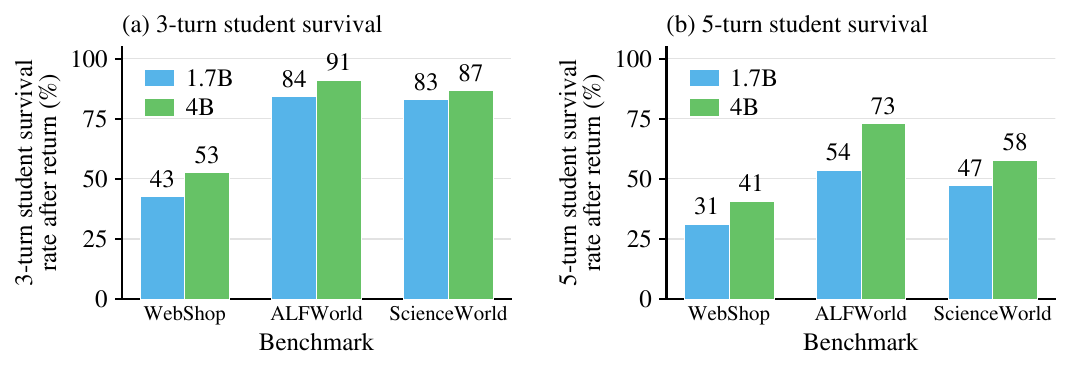}
    \caption{Student survival rates after the teacher returns control
    during DASH-OPD training at \textbf{(a)} three-turn and \textbf{(b)}
    five-turn horizons. Survival means that the student executes the full
    horizon without another teacher intervention or completes the task
    successfully before the horizon. The substantial post-return survival
    rates suggest that teacher intervention helps students continue independently.}
    \label{fig:switch-anatomy-app}
\end{figure}

We sample rollouts every 50 training steps from the six DASH-OPD training
runs in \cref{tab:main-results-all-envs} and count the consecutive student
or teacher turns before each executor switch, as shown in \cref{fig:preceding-run-lengths}. The sample contains 386
student-to-teacher and 436 teacher-to-student switches. Before teacher intervention, the
student remains in control for an average of 4.43 turns at 1.7B and 5.10
turns at 4B. Three consecutive student turns precede 41.75\% and 39.58\%
of student-to-teacher switches, respectively. Before returning control to
the student, the teacher remains in control for an average of 4.09 turns
at 1.7B and 3.61 turns at 4B. Two
consecutive teacher turns precede 45.45\% and 53.24\% of
teacher-to-student switches, respectively. At 4B, the student
remains in control longer and the teacher remains in control for fewer
turns than at 1.7B. Most switches occur after multiple
consecutive turns under one executor, providing further evidence of
hysteretic switching.

\section{Student survival after the teacher returns control}
\label{app:student-survival-after-resuming-control}

\Cref{fig:switch-anatomy-app} complements the mean post-return student
spans in \cref{fig:switch-efficiency-anatomy}(b) with survival rates at two
horizons. At a three-turn horizon, the 1.7B and 4B rates are 84\% and 91\% on
ALFWorld, 83\% and 87\% on ScienceWorld, and 43\% and 53\% on WebShop.
At a five-turn horizon, they are 54\% and 73\%, 47\% and 58\%, and 31\% and
41\%, respectively. WebShop's 15-turn episode limit, versus 30 turns on
the other benchmarks, can shorten its mean post-return student
spans. At both horizons and on every benchmark, survival is higher for the 4B
student, consistent with its longer mean post-return spans. Overall, the high
post-return survival rates, especially at three turns, further support the
effectiveness of teacher intervention in sustaining student autonomy.

\section{Robustness of action-only discrepancy}
\label{app:action-extraction-robustness}

If the action extraction for \cref{eq:student-ratio,eq:teacher-ratio} fails, the rollout is
discarded without updating the controller, and training continues
with other samples. In ALFWorld, such failures affect only
0.60\% and 0.15\% of trajectories for the 1.7B and 4B students,
respectively. After excluding failed rollouts, the 1.7B run collects 3.92 times the number of samples required for training, and the 4B run collects 3.91 times the required number, leaving substantial sample surpluses in both runs. The runs on ScienceWorld and WebShop also collect more samples than required (3.18/2.87 times for 1.7B/4B on ScienceWorld and 1.20/1.34 times for 1.7B/4B on WebShop), leaving a surplus to accommodate rollout failures.
Moreover, early in training, the high teacher-start probability in \cref{eq:teacher-start} reduces
reliance on student-generated outputs, while full-response
distillation helps the student learn the action format.
Each DASH-OPD training run completes without termination due to action extraction failures.

\section{Handling sudden increases in task difficulty}
\label{app:sudden-difficulty}

DASH-OPD supports teacher intervention when task difficulty suddenly increases after the student resumes control. Such increases can potentially be detected through a large teacher--student discrepancy or high student entropy. As described in \cref{sec:method}, a sufficiently large teacher--student discrepancy on the first resumed student turn can immediately trigger teacher intervention on the next turn. We also evaluate an entropy-based fallback that discards the first resumed student response and retains teacher control if the response's mean token entropy exceeds $2$. This fallback, however, is never triggered in our experiments.

\section{Prompt templates}
\label{app:prompt-templates}

All methods use the same environment-specific user-prompt templates in
training and evaluation. Given the same interaction history, the teacher and 
student receive identical prompts and use the same action parser. The full
user--assistant conversation remains in context throughout the episode.

The first three turns have no explicit history summary. From the fourth turn onward, \texttt{\{action\_history\}} contains the two most recent
observation--action pairs, each formatted as
\texttt{[Observation $i$: \textquotesingle{}$o_i$\textquotesingle{}, Action $i$: \textquotesingle{}$a_i$\textquotesingle{}]}.

The prompts ask for reasoning followed by an action. The environment action must
appear between \texttt{<action>} and \texttt{</action>} for parsing. The
boxes below reproduce the templates used in the experiments.

\paragraph{ALFWorld.}
In ALFWorld, the agent navigates and manipulates objects to complete
household tasks. Each turn supplies the current observation and admissible
actions. For the first three turns, the prompt has no separate task or
history field: the task appears in the initial observation, which remains
in the conversation history.

\begin{promptbox}
You are an expert agent operating in the ALFRED Embodied
Environment. Your current observation is:
{current_observation} Your admissible actions of the current
situation are: [{admissible_actions}].

Now it's your turn to take an action. You should first reason
about the current situation. Once you've finished your
reasoning, you should choose the best admissible action for
the current step and present it within <action> </action>
tags. Do not output any other text besides your reasoning and
the action.
\end{promptbox}

From the fourth turn onward, the prompt also includes the task description, the step count,
and the two most recent observation--action pairs:

\begin{promptbox}
You are an expert agent operating in the ALFRED Embodied
Environment. Your task is to: {task_description} Prior to this
step, you have already taken {step_count} step(s). Below are
the most recent {history_length} observations and the
corresponding actions you took: {action_history} You are now
at step {current_step} and your current observation is:
{current_observation} Your admissible actions of the current
situation are: [{admissible_actions}].

Now it's your turn to take an action. You should first reason
about the current situation. Once you've finished your
reasoning, you should choose the best admissible action for
the current step and present it within <action> </action>
tags. Do not output any other text besides your reasoning and
the action.
\end{promptbox}

\paragraph{ScienceWorld.}
In ScienceWorld, the agent conducts experiments over multiple turns. The
prompt lists action templates and available objects separately. The agent
combines them into a valid action. The first three turns use the following
template without an explicit history summary:

\begin{promptbox}
You are an expert agent operating in the ScienceWorld text
environment. {task_description} Your current observation is:
{current_observation} Available action commands:
[{action_templates}] Available objects you can interact with:
[{objects}]

Now it's your turn to take an action. Combine an action
command with appropriate object(s) to form a valid action. You
should first reason about the current situation. Once you've
finished your reasoning, you should choose the best valid
action for the current step and present it within <action>
</action> tags. Do not output any other text besides your
reasoning and the action.
\end{promptbox}

From the fourth turn onward, the prompt also includes the step count and the two most recent
observation--action pairs:

\begin{promptbox}
You are an expert agent operating in the ScienceWorld text
environment. {task_description} Prior to this step, you have
already taken {step_count} step(s). Below are the most recent
{history_length} observations and the corresponding actions
you took: {action_history} You are now at step {current_step}
and your current observation is: {current_observation}
Available action commands: [{action_templates}] Available
objects you can interact with: [{objects}]

Now it's your turn to take an action. Combine an action
command with appropriate object(s) to form a valid action. You
should first reason about the current situation. Once you've
finished your reasoning, you should choose the best valid
action for the current step and present it within <action>
</action> tags. Do not output any other text besides your
reasoning and the action.
\end{promptbox}

\paragraph{WebShop.}
In WebShop, the agent navigates an e-commerce interface to satisfy a
natural-language shopping request. Available actions depend on the current
page. They primarily include \texttt{search[query]} when a search bar is
present and \texttt{click[button\_name]}, whose argument must match a
clickable element. The first three turns use the following template:

\begin{promptbox}
You are an expert autonomous agent operating in the WebShop
e-commerce environment. Your task is to: {task_description}.
Your current observation is: {current_observation}. Your
admissible actions of the current situation are: [
{available_actions} ].

Now it's your turn to take one action for the current step.
You should first reason about the current situation, then
think which admissible action best advances the shopping goal.
Once you've finished your reasoning, you should choose the
best admissible action for the current step and present it
within <action> </action> tags. Do not output any other text
besides your reasoning and the action.
\end{promptbox}

From the fourth turn onward, the prompt also includes the step count and the two most recent
observation--action pairs:

\begin{promptbox}
You are an expert autonomous agent operating in the WebShop
e-commerce environment. Your task is to: {task_description}.
Prior to this step, you have already taken {step_count}
step(s). Below are the most recent {history_length}
observations and the corresponding actions you took:
{action_history} You are now at step {current_step} and your
current observation is: {current_observation}. Your admissible
actions of the current situation are: [ {available_actions} ].

Now it's your turn to take one action for the current step.
You should first reason about the current situation, then
think which admissible action best advances the shopping goal.
Once you've finished your reasoning, you should choose the
best admissible action for the current step and present it
within <action> </action> tags. Do not output any other text
besides your reasoning and the action.
\end{promptbox}

\section{Detailed training and evaluation settings}
\label{app:full-configurations}

\paragraph{Settings shared by all methods.}
For each benchmark and student scale, all six methods use the same
teacher--student pair, training framework, interaction protocol, and
evaluation protocol. \Cref{tab:shared-configurations} lists these settings;
most follow the official TCOD implementation.

\begin{table}[h]
    \centering
    \caption{Training and evaluation settings shared by all six methods at
    both model scales. \enquote{No sampling truncation} denotes
    top-$p=1$, top-$k=-1$, and min-$p=0$.}
    \label{tab:shared-configurations}
    \scriptsize
    \setlength{\tabcolsep}{3.5pt}
    \renewcommand{\arraystretch}{1.12}
    \begin{tabularx}{\textwidth}{@{}>{\raggedright\arraybackslash}p{0.25\textwidth}
        *{3}{>{\centering\arraybackslash}X}@{}}
        \toprule
        Setting & WebShop & ALFWorld & ScienceWorld \\
        \midrule
        Student model / frozen teacher
            & \multicolumn{3}{>{\centering\arraybackslash}p{0.71\textwidth}}{Qwen3-1.7B or Qwen3-4B / Qwen3-30B-A3B} \\
        Training task pool size
            & 4,096
            & 3,553
            & 2,294 \\
        Optimizer updates
            & 150 & 250 & 250 \\
        Rollout / update batch size
            & \multicolumn{3}{>{\centering\arraybackslash}p{0.71\textwidth}}{16 / 64} \\
        Rollout staleness
            & \multicolumn{3}{>{\centering\arraybackslash}p{0.71\textwidth}}{At most two student policy versions behind the current version} \\
        Optimizer
            & \multicolumn{3}{>{\centering\arraybackslash}p{0.71\textwidth}}{AdamW; $\mathrm{lr}=10^{-6}$; $(\beta_1,\beta_2)=(0.9,0.999)$; weight decay $0.01$} \\
        Learning-rate schedule / warm-up
            & \multicolumn{3}{>{\centering\arraybackslash}p{0.71\textwidth}}{Constant / none} \\
        Gradient clipping
            & \multicolumn{3}{>{\centering\arraybackslash}p{0.71\textwidth}}{Global gradient norm capped at $1.0$} \\
        Additional RL components
            & \multicolumn{3}{>{\centering\arraybackslash}p{0.71\textwidth}}{No critic, reference-policy KL penalty, or entropy bonus} \\
        Training decoding
            & \multicolumn{3}{>{\centering\arraybackslash}p{0.71\textwidth}}{Temperature $1.0$; no sampling truncation; at most 512 new tokens per turn} \\
        Maximum prompt length
            & \multicolumn{3}{>{\centering\arraybackslash}p{0.71\textwidth}}{10,240 tokens} \\
        Maximum turns per task $H_{\max}$
            & $15$ & $30$ & $30$ \\
        Interaction context
            & \multicolumn{3}{>{\centering\arraybackslash}p{0.71\textwidth}}{Full conversation history; two latest observation--action pairs summarized from turn 4} \\
        Model thinking mode
            & \multicolumn{3}{>{\centering\arraybackslash}p{0.71\textwidth}}{Qwen3 thinking mode disabled} \\
        Random seed
            & \multicolumn{3}{>{\centering\arraybackslash}p{0.71\textwidth}}{42 for all stochastic components in training and evaluation} \\
        Training precision
            & \multicolumn{3}{>{\centering\arraybackslash}p{0.71\textwidth}}{bfloat16} \\
        Memory settings
            & \multicolumn{3}{>{\centering\arraybackslash}p{0.71\textwidth}}{Dynamic batching; 16,384 tokens per GPU; Ulysses sequence parallelism \cite{ulysses} $=2$} \\
        Hardware
            & \multicolumn{3}{>{\centering\arraybackslash}p{0.71\textwidth}}{One node with eight NVIDIA A100 GPUs, each with 80\,GB of memory} \\
        Evaluated policy
            & \multicolumn{3}{>{\centering\arraybackslash}p{0.71\textwidth}}{Final student checkpoint; no teacher or rollout controller} \\
        Evaluation set
            & 100 tasks
            & 140 IID + 134 OOD tasks
            & 1,308 tasks \\
        Evaluation rollouts
            & \multicolumn{3}{>{\centering\arraybackslash}p{0.71\textwidth}}{One rollout per task} \\
        Evaluation batch size
            & \multicolumn{3}{>{\centering\arraybackslash}p{0.71\textwidth}}{16 tasks} \\
        Evaluation decoding
            & \multicolumn{3}{>{\centering\arraybackslash}p{0.71\textwidth}}{Temperature $0.4$; no sampling truncation; at most 4,096 new tokens per turn} \\
        Reported metrics
            & Mean score; SR; mean turns
            & SR and mean turns
            & Mean score; SR; mean turns \\
        \bottomrule
    \end{tabularx}
\end{table}

\paragraph{Baseline-specific settings and behavior.}
\Cref{tab:baseline-configurations} specifies each baseline's hyperparameters,
rollout rules, and training objective. We follow the original papers or
official implementations where applicable. Settings that require adaptation
to this experimental setup receive the same tuning budget as DASH-OPD.

\begin{table}[t]
    \centering
    \caption{Baseline-specific settings for both student model scales.
    Sampled reverse-KL losses use PPO clipping with $\epsilon=0.2$ and a
    dual-clip coefficient of $3$. TCOD and Guided-OPD advance their curricula
    once per rollout batch; TurnOPD numbers updates using the rollout student
    version plus one. NLL denotes negative log-likelihood, and EMA denotes
    exponential moving average.}
    \label{tab:baseline-configurations}
    \scriptsize
    \setlength{\tabcolsep}{2.5pt}
    \renewcommand{\arraystretch}{1.10}
    \begin{tabularx}{\textwidth}{@{}>{\raggedright\arraybackslash}p{0.13\textwidth}
        >{\raggedright\arraybackslash}p{0.23\textwidth}
        *{3}{>{\centering\arraybackslash}X}@{}}
        \toprule
        Baseline & Setting & WebShop & ALFWorld & ScienceWorld \\
        \midrule
        \multirow{2}{*}{Vanilla OPD}
            & Rollout executor
            & \multicolumn{3}{>{\centering\arraybackslash}p{0.60\textwidth}}{Student at every turn} \\
            & Objective
            & \multicolumn{3}{>{\centering\arraybackslash}p{0.60\textwidth}}{Sampled reverse KL on student-generated tokens, scored by the teacher} \\
        \midrule
        \multirow{5}{*}{AdaSwitch}
            & Token-level rollout rule
            & \multicolumn{3}{>{\centering\arraybackslash}p{0.60\textwidth}}{Begin each turn with the student; after a switch, the teacher completes that turn} \\
            & Switching statistic
            & \multicolumn{3}{>{\centering\arraybackslash}p{0.60\textwidth}}{Truncated forward-KL proxy: renormalize each model's top-1,024 token probabilities, sum over tokens in both top-1,024 sets, and clamp negative values to zero} \\
            & Window $L$ / multiplier $k$
            & $10$ / $3$ & $10$ / $3$ & $10$ / $3$ \\
            & Switching criterion
            & \multicolumn{3}{>{\centering\arraybackslash}p{0.60\textwidth}}{Switch when the current divergence exceeds $k$ times the mean of the previous $L$ values; require a full window before switching} \\
            & Objective
            & \multicolumn{3}{>{\centering\arraybackslash}p{0.60\textwidth}}{Token-averaged weighted NLL: weights $\pi_T/\pi_{\bar\theta}$ on student-generated tokens and $1$ on teacher-generated tokens; no PPO clipping} \\
        \midrule
        \multirow{5}{*}{TCOD}
            & Reported result
            & \multicolumn{3}{>{\centering\arraybackslash}p{0.60\textwidth}}{Average of backward-to-forward (B2F) and forward-to-backward (F2B) results} \\
            & B2F / F2B interval (batches)
            & 4 / 4 & 5 / 6 & 5 / 6 \\
            & B2F curriculum
            & \multicolumn{3}{>{\centering\arraybackslash}p{0.60\textwidth}}{Replay an expert prefix, then let the student continue; shorten the prefix by one action at each B2F interval until no expert actions remain} \\
            & F2B curriculum
            & \multicolumn{3}{>{\centering\arraybackslash}p{0.60\textwidth}}{Student horizon $\min\{H_{\max},1+\lfloor s/\Delta_{\mathrm{F2B}}\rfloor\}$, where $s$ is the zero-based rollout batch index and $\Delta_{\mathrm{F2B}}$ is the F2B interval} \\
            & Objective
            & \multicolumn{3}{>{\centering\arraybackslash}p{0.60\textwidth}}{Sampled reverse KL on student-generated tokens} \\
        \midrule
        \multirow{6}{*}{Guided-OPD}
            & Executor selection
            & \multicolumn{3}{>{\centering\arraybackslash}p{0.60\textwidth}}{Choose the teacher or student independently at each turn (seed 42)} \\
            & Teacher-selection decay
            & \multicolumn{3}{>{\centering\arraybackslash}p{0.60\textwidth}}{Cosine decay from $1$ to $0$} \\
            & Decay duration (batches)
            & 120 & 200 & 200 \\
            & Final student-only batches
            & 30 & 50 & 50 \\
            & Executor-specific objective
            & \multicolumn{3}{>{\centering\arraybackslash}p{0.60\textwidth}}{Sampled reverse KL on student turns; NLL on teacher turns (sampled forward KL up to a constant)} \\
            & Sequence masking / fallback
            & \multicolumn{3}{>{\centering\arraybackslash}p{0.60\textwidth}}{Sequence masking and policy-gradient fallback disabled} \\
        \midrule
        \multirow{11}{*}{TurnOPD}
            & Depth: min. / requested max. / effective max.
            & 2 / 50 / 15 & 2 / 50 / 30 & 2 / 50 / 30 \\
            & Probe schedule
            & \multicolumn{3}{>{\centering\arraybackslash}p{0.60\textwidth}}{Start the first probe during updates 1--3; start each later probe after 8 updates have elapsed since the preceding probe finished} \\
            & Probe batch size
            & \multicolumn{3}{>{\centering\arraybackslash}p{0.60\textwidth}}{64 full-depth trajectories ($4\times16$)} \\
            & Controller signal
            & \multicolumn{3}{>{\centering\arraybackslash}p{0.60\textwidth}}{Reverse KL on the teacher's top-50 tokens, with both distributions renormalized} \\
            & Coverage horizon
            & \multicolumn{3}{>{\centering\arraybackslash}p{0.60\textwidth}}{Nearest-rank 80th percentile of successful completion turns; retain the previous value with fewer than 8 successes (initial value $0$)} \\
            & EMA weight on new horizon
            & $0.30$ & $0.30$ & $0.30$ \\
            & Adaptive rollout depth
            & \multicolumn{3}{>{\centering\arraybackslash}p{0.60\textwidth}}{Initialize $\bar H$ to the effective maximum; use EMA to track the larger of the zero-based KL-mass and coverage horizons; clip $\lfloor\bar H+0.5\rfloor+1$ to the depth bounds} \\
            & KL-mass normalization $\epsilon$
            & $10^{-8}$ & $10^{-8}$ & $10^{-8}$ \\
            & Turn-normalization blend
            & Linear $0\!\rightarrow\!1$; 150 updates
            & Linear $0\!\rightarrow\!1$; 250 updates
            & Linear $0\!\rightarrow\!1$; 250 updates \\
            & Minimum support per turn
            & \multicolumn{3}{>{\centering\arraybackslash}p{0.60\textwidth}}{$\max\{8,\lceil0.15N\rceil\}$ trajectories, where $N$ is the number returned in the rollout batch; use uniform token weights if no turn meets this threshold} \\
            & Objective
            & \multicolumn{3}{>{\centering\arraybackslash}p{0.60\textwidth}}{Sampled reverse KL, progressively shifting from uniform token weighting to equal weighting of valid turns} \\
        \bottomrule
    \end{tabularx}
\end{table}

\section{Limitations and future work}
Our experiments cover three text-based benchmarks and one teacher--student model family. Evaluating DASH-OPD across other modalities and model families would clarify whether its performance gains generalize beyond the current experimental settings. In addition, DASH-OPD uses a single fixed teacher during training. Future work could explore multiple domain-specialized teachers and assess whether they improve performance across diverse tasks.

%%%%%%%%%%%%%%%%%%%%%%%%%%%%%%%%%%%%%%%%%%%%%%%%%%%%%%%%%%%%

\newpage

\end{document}